\documentclass[letterpaper]{article} 
\usepackage[preprint]{aaai2027}  
\usepackage[hyphens]{url}  
\usepackage{graphicx} 
\usepackage{natbib}  
\usepackage{caption} 
\usepackage{booktabs}

\usepackage{amsmath}
\usepackage{amssymb}
\usepackage{amsthm}
\usepackage{enumitem}

\theoremstyle{plain} %
\newtheorem{theorem}{Theorem}[section]
\newtheorem{proposition}{Proposition} %
\newtheorem{lemma}[theorem]{Lemma}

\theoremstyle{definition} %

\theoremstyle{remark} %

\title{Post-Hoc Tail Calibration of Probabilistic Predictions\\Above an Operational Threshold}
\author{
    Novin Shahroudi,
    Meelis Kull
}
\affiliations {
    Institute of Computer Science\\
    University of Tartu, Estonia\\
    novin.shahroudi@ut.ee
}

\begin{document}
\maketitle

\begin{abstract}

Many operational decisions depend on whether an outcome exceeds a capacity, safety, or risk threshold, making reliable characterization of the tails of predictive distributions consequential.
Common notions of reliability such as marginal quantile calibration and calibration of the probability integral transform (PIT) average over cases, allowing threshold-specific errors to cancel across heterogeneous predictions.
We propose a model-agnostic post-hoc calibration method for models that produce univariate predictive distributions when a threshold of interest is known in advance. Using held-out calibration data, it separately recalibrates threshold exceedance probabilities and excess-PIT values, which represent the percentiles of observed exceedances within their predicted upper tails, before recombining the corrections into a complete predictive distribution. We show that the construction defines a valid distribution and, under population recalibration maps and stated regularity conditions, attains occurrence and excess-PIT calibration at the chosen threshold while preserving the normalized distribution below it. Controlled synthetic experiments examine targeted corrections and finite-sample estimation effects. Across eight UCI datasets, PostTail improves aggregate fixed-threshold calibration relative to unmodified predictions and global PIT recalibration, with trade-offs in proper scores and global calibration.

\end{abstract}

\section{Introduction}
\label{sec:introduction}

Probabilistic predictions are used to support decisions under uncertainty, but in many applications the most consequential decisions are driven by the upper tail of the predictive distribution. Operators often need to know whether a system will enter a rare high-impact regime, e.g., demand will exceed capacity, or whether weather variables will cross safety thresholds, or losses will exceed a risk limit. In these settings, reliability near an operational threshold can matter more than average performance over the full distribution. %

\begin{figure}[th]
    \centering
    \includegraphics[width=1.0\linewidth]{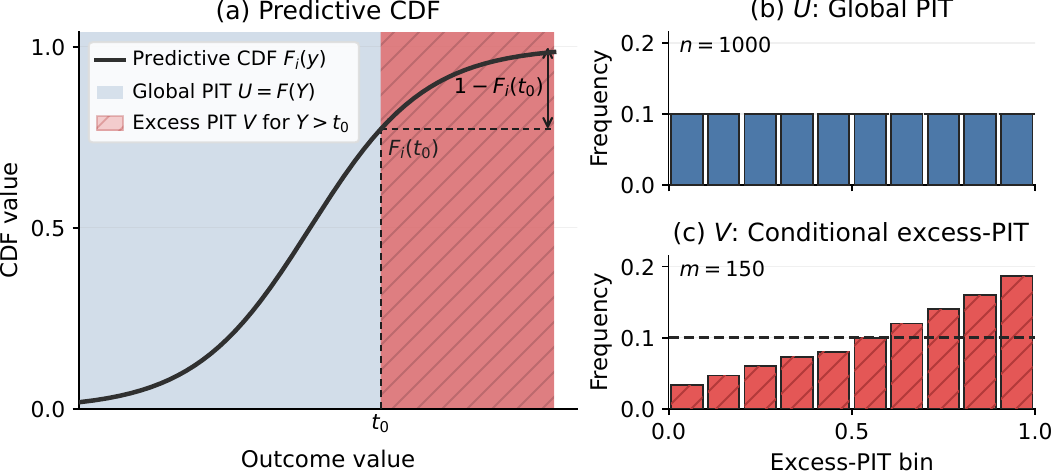}
    \caption{
    Schematic PIT diagnostics for calibration pairs $\{(F_i,y_i)\}_{i=1}^n$, drawn with $n\!=\!1000$ observations and $m\!=\!150$ exceedances above a fixed threshold $t_0$, shows that a globally uniform PIT histogram can mask miscalibration within the predicted tail. (a) shows the full CDF scale used by global PIT values $u_i\!=\!F_i(y_i)$ and the upper-tail segment used after conditioning on $y_i>t_0$. (b) bins all $u_i$ and estimates $\Pr(U\in B_k)$ by $n_k/n$. (c) bins excess-PIT values $v_i\!=\!(u_i-F_i(t_0))/(1-F_i(t_0))$ for exceedances and estimates $\Pr(V\in B_k\mid Y>t_0)$ by $m_k/m$. With ten equal bins, uniform calibration gives target mass $0.1$ per bin (dashed lines). For example, $n\!=\!100$ would imply about 10 observations per global bin, while $m\!=\!20$ would imply about 2 exceedances per tail bin. %
    }
    \label{fig:firstpage_illustration}
    \vspace{-0.5em}
\end{figure}

A standard requirement for probabilistic predictions is calibration, an umbrella term for several related notions of agreement between predicted probabilities and observed frequencies. Classical probabilistic prediction evaluation emphasizes calibration and sharpness, often through PIT-based diagnostics~\cite{gneiting2007probabilistic} and proper scoring rules~\cite{gneiting2007strictly}. For a continuous random outcome $Y$ with predictive cumulative distribution function (CDF) $F$, the PIT $U=F(Y)$ is the outcome's predicted percentile. If $F$ is correctly specified, $U$ is uniform on $[0,1]$, motivating a uniform histogram as a global calibration diagnostic. Post-hoc recalibration methods further use held-out calibration data to transform predictive distributions after training, including quantile and PIT-map approaches~\cite{kuleshov2018accurate, zelikman2020crude, dheur2023large, qian2025regression}. However, global calibration criteria can average over the body of the distribution and obscure threshold-specific failures. Figure~\ref{fig:firstpage_illustration} previews this distinction. The PIT histogram over all prediction--outcome pairs appears uniform, yet the excess-PIT values conditional on $Y>t_0$ are nonuniform, revealing miscalibration within the predicted upper tail.

Recent work has clarified why this distinction matters. \citet{sahoo2021reliable} show that average calibration is not sufficient for reliable threshold-decision loss estimation and introduce threshold calibration for decisions based on crossing cutoffs. \citet{allen2025tail} develop a theory of probabilistic tail calibration and decompose tail reliability into two components: occurrence, concerning agreement between exceedance frequencies and predicted exceedance probabilities, and severity, concerning uniformity of excess-PIT values conditional on exceedance.

Existing methods address parts of this problem, but leave a practical gap. \citet{wessel2026enforcing} study how to enforce tail calibration during model training, for example by modifying training losses. This is valuable when the predictive model can be retrained, but many deployed prediction systems are fixed, proprietary, computationally expensive, or only available through samples or CDF queries. Other post-hoc approaches reshape predictive distributions through global PIT maps or more general transport maps.
The LADaR framework learns covariate-dependent probability-to-probability maps that diagnose and recalibrate full conditional predictive distributions~\cite{dey2025towards}. Recent semiparametric diagnostic transport maps extend this framework to settings with limited calibration data and explicit emphasis on tail events~\cite{cucuzzella2026trustworthy}, while score-based methods learn global or prediction-dependent PIT maps~\cite{qian2025regression}.
Model-specific approaches such as tcGP instead target lower-tail calibration for Gaussian-process Bayesian optimization~\cite{pion2026goal}. These methods offer flexible global predictive-CDF reshaping, whereas our target is narrower. We study post-hoc calibration at and above a prespecified outcome threshold while preserving the normalized prediction body, limiting the intervention to the targeted tail and leaving relative probabilities among subthreshold outcomes unchanged.

Given a fixed operational threshold and an independent calibration sample, the goal is to adjust the prediction's reliability near that threshold without retraining the base model. High-impact weather provides a concrete application, namely, probabilistic warning systems combine event likelihood with escalating severity levels for hazards such as extreme heat, heavy rainfall, and damaging winds~\cite{allen2023weighted,taggart2025warnings}. The crossing probability can inform whether a warning is needed, while the predicted conditional tail characterizes how severe the event may be once the threshold is crossed.

\vspace{0.5em}
\noindent
Therefore, our contributions are as follows:
\begin{itemize}[nosep]
    \item We introduce a model-agnostic fixed-threshold post-hoc recalibration method that recalibrates exceedance probabilities and excess-PIT values of pretrained univariate predictive CDFs with separate calibration maps, then combines the two maps to form new CDFs.
    \item We show that the method returns valid distributions and the population maps attain occurrence and excess-PIT calibration while preserving the normalized prediction body. %
    \item We evaluate the method empirically against unmodified predictions and global PIT recalibration using synthetic and real-data experiments.
\end{itemize}

\vspace{0.5em}
Section~\ref{sec:background} introduces notation and calibration targets. Sections~\ref{sec:method} and~\ref{sec:estimation} present the population maps and their finite-sample estimation. Section~\ref{sec:experiments} reports synthetic and real-data experiments, and Section~\ref{sec:conclusion} discusses conclusions.

\section{Background}
\label{sec:background}

We study post-hoc calibration for univariate predictive cumulative distribution functions (CDFs). Let \(Y\in\mathbb R\) denote the target random variable and let \(F\) denote the predictive CDF issued for \(Y\). We treat \((F,Y)\) as a random prediction--outcome pair. We write \(y\) for a possible realization of \(Y\). The predictive CDF \(F\) may be produced by any pretrained probabilistic model. The post-hoc procedures considered here only require access to CDF values, or to a sample representation from which CDF values can be estimated.

\subsection{Probabilistic Calibration and PIT Recalibration}
For a continuous predictive CDF, the probability integral transform (PIT) of the random prediction--outcome pair is
\[
U=F(Y).
\]
For a realized prediction--outcome pair, the corresponding PIT value is \(u=F(y)\). If \(F\) is the true conditional distribution of \(Y\), then \(U\) is uniformly distributed on \([0,1]\). PIT histograms and probability--probability plots therefore provide standard diagnostics for probabilistic calibration~\cite{diebold1997evaluating, gneiting2007probabilistic}. Global PIT calibration asks whether these probability values are uniform after averaging over prediction instances. Post-hoc PIT recalibration methods estimate a monotone probability-scale map from held-out calibration data and apply it to the predictive CDF, as in quantile recalibration or Kuleshov-style isotonic recalibration~\cite{kuleshov2018accurate, qian2025regression}.

Global PIT calibration is a pooled, marginal probability-scale target. 
It can be useful when pooled PIT uniformity is the relevant target, but it does not necessarily achieve calibration within a particular operational region.
A single global map may correct central and tail PIT errors together, and heterogeneous errors can cancel when all prediction instances are pooled. Proper scoring rules such as Continuous Ranked Probability Score (CRPS) summarize predictive accuracy and calibration jointly~\cite{gneiting2007strictly}, threshold-weighted variants such as upper-tail twCRPS emphasize a region of the outcome space~\cite{allen2023evaluating,allen2024weighted}. These scores are useful companions to calibration diagnostics, but they do not by themselves define the fixed-threshold calibration target studied below.

\subsection{Fixed Outcome Threshold} 

Let \(t\in\mathbb R\) be a fixed upper operational threshold. The threshold \(t\) is chosen before fitting the post-hoc recalibrator and represents the tail region of practical interest. Define the exceedance-probability functional at \(t\) by
\[
p_t(F) = 1 - F(t).
\]
For brevity, when the threshold is clear from context, we write \(P=p_t(F)\).
The observed exceedance indicator is
\[
E = \mathbf 1\{Y>t\}.
\]
For outcomes above the threshold, define the predictive conditional upper-tail, or excess, CDF
\[
F_t^+(y)
=
\frac{F(y)-F(t)}{1-F(t)},
\qquad y>t,
\]
whenever \(0\!<\!F(t)\!<\!1\). The superscript \(+\) indicates conditioning on the upper tail above \(t\). The corresponding excess PIT is
\begin{equation}
Z
=
F_t^+(Y)
=
\frac{F(Y)-F(t)}{1-F(t)},
\qquad \text{on } \{Y>t\}.
\label{eq:excess_pit}
\end{equation}

Thus \(P\) denotes the random upper-tail probability across prediction instances, while \(Z\), defined on exceedances, denotes the probability-scale position, or conditional percentile, of the outcome within its predicted upper tail. We have presented the above notation for upper-tail exceedances, but a lower-tail version can be obtained by replacing \(Y>t\) with \(Y<t\), using \(F(t)\) as the predicted lower-tail probability, and defining the analogous conditional percentile within the lower tail. We do not address simultaneous two-sided or multi-threshold calibration in this paper.

\subsection{Tail Calibration at a Fixed Threshold}

Following the occurrence--severity decomposition of probabilistic tail calibration~\cite{allen2025tail,wessel2026enforcing}, reliability at \(t\) has two components.
The first is occurrence calibration. The weak marginal occurrence condition is
\[
\Pr(Y>t)=\mathbb E[P].
\]
A stronger probability-reliability form is
\begin{equation}
\Pr(Y>t\mid P)=P.
\label{eq:occurrence_calibration}
\end{equation}
Equation~\eqref{eq:occurrence_calibration} is the occurrence-calibration analogue of reliability for binary probability predictions. It asks whether cases assigned the same exceedance probability exceed the threshold at the corresponding empirical frequency.\\
The second component is severity calibration, defined through excess-PIT uniformity:
\begin{equation}
Z \mid Y>t \sim \operatorname{Unif}(0,1).
\label{eq:excess_severity_calibration}
\end{equation}
Here \(Z\) is the excess-PIT value defined in Equation~\eqref{eq:excess_pit}, or equivalently the conditional percentile of an outcome within its predicted upper tail. Equation~\eqref{eq:excess_severity_calibration} asks whether observed exceedances are uniformly distributed across the predicted conditional tail, rather than systematically appearing too low or too high within that tail. \\
The marginal occurrence condition above and the excess-severity condition in Equation~\eqref{eq:excess_severity_calibration} imply the marginal combined fixed-threshold condition by the multiplication rule:
\begin{equation}
\begin{aligned}
\Pr(Z\le u,Y>t)
&=
\Pr(Z\le u\mid Y>t)\Pr(Y>t)\\
&=
u\,\mathbb E[P],
\qquad u\in[0,1].
\end{aligned}
\label{eq:combined_fixed_threshold_calibration}
\end{equation}
Equivalently, define the combined fixed-threshold calibration curve
\begin{equation}
R_t(u)
=
\frac{\Pr(Z\le u, Y>t)}{\mathbb E[P]},
\qquad u\in[0,1],
\label{eq:combined_calibration_curve}
\end{equation}
which rewrites Equation~\eqref{eq:combined_fixed_threshold_calibration} as the target \(R_t(u)=u\). 

In empirical evaluations, let \(\widehat R_t\) denote the estimate of \(R_t\) computed from held-out data. This estimated curve can be displayed in a probability--probability plot against the identity line or summarized by the measure of tail miscalibration (TMCB) introduced by~\cite{wessel2026enforcing}, defined as its integrated absolute deviation from the identity target:
\[
\operatorname{TMCB}_t
=
\int_0^1 |\widehat R_t(u)-u|\,du.
\]

The empirical combined curve and its TMCB summary are pooled, fixed-threshold diagnostics. They assess the combined occurrence and excess-severity condition after averaging across prediction instances. %
They do not establish calibration conditional on prediction features or simultaneous calibration across multiple thresholds. %
This fixed-threshold target differs from the asymptotic tail-calibration definition of~\cite{allen2025tail}, where analogous conditions are studied as the threshold approaches the endpoint of the outcome distribution. It is also narrower than threshold calibration for broad families of threshold-based decisions~\cite{sahoo2021reliable}. The target here is deliberately operational where the threshold is fixed in advance because it is the cutoff at which the prediction will be used.

\section{Post-hoc Tail Recalibration} 
\label{sec:method}
This section introduces the fixed-threshold recalibration target and the post-hoc operator in terms of population recalibration maps. Their finite-sample estimation is described in Section~\ref{sec:estimation}. 
The proposed recalibrator uses two one-dimensional population maps, namely an occurrence map \(r\) for tail probabilities and a severity map \(s\) for excess-PIT values.

\subsubsection{Occurrence Map.} 
The occurrence component uses the classical reliability function for a binary probability prediction, namely the conditional event probability given the issued probability~\cite{broecker2009reliability}. We call this function the occurrence map because \(E=1\) indicates that the threshold is exceeded. Its population form is
\begin{equation}
r(p) = \Pr(E=1\mid P=p),
\qquad p\in[0,1].
\label{eq:occurrence_map}
\end{equation}
For a value \(p\) attained by \(P\), \(r(p)\) is the population exceedance probability among prediction instances assigned probability \(p\). Thus, \(r(p)>p\) indicates that the original prediction underestimates the exceedance probability, whereas \(r(p)<p\) indicates that it overestimates it. The recalibrated exceedance probability is
\(
P' = r(P).
\)
This conditional-probability transformation is itself occurrence calibrated at the population level~\cite{ranjan2010combining}. If the original occurrence probabilities are calibrated, then \(r(p)=p\) on the support of \(P\), and this step reduces to the identity.

\subsubsection{Excess-Severity Map.} 
Among exceedances, the population excess-PIT CDF is
\[
H(u) = \Pr(Z\le u\mid E=1).
\]
The population severity map is
\begin{equation}
s(u)=H(u).
\label{eq:severity_map}
\end{equation}
Thus \(H\) describes the population distribution of the raw excess-PIT values, while Equation~\eqref{eq:severity_map} gives the population map that would uniformize those values.
The recalibrated conditional excess CDF is
\[
F_t^{+\prime}(y)
=
s(F_t^+(y)),
\qquad y>t.
\]

If the base prediction is already well calibrated in the tail, then \(r(p)\approx p\) and \(s(u)\approx u\), so the operator remains close to the identity.

\subsubsection{Recombined Predictive CDF.} 
The full recalibrated CDF is obtained by preserving the normalized body distribution below the threshold, replacing the tail mass by \(P'\), and applying \(s\) inside the conditional tail:
\begin{equation}
F'(y)=
\begin{cases}
(1-P')\dfrac{F(y)}{F(t)},
& y\le t,\\[10pt]
1-P'
+P's\!\left(
\dfrac{F(y)-F(t)}{1-F(t)}
\right),
& y>t.
\end{cases}
\label{eq:posttail_population_cdf}
\end{equation}
This construction changes the total mass assigned above \(t\), but it leaves the normalized shape of the prediction below \(t\) unchanged:
\[
\frac{F'(y)}{F'(t)}
=
\frac{F(y)}{F(t)},
\qquad y\le t.
\]
It also preserves the prediction ordering within the conditional tail whenever \(s\) is nondecreasing, and strictly preserves ordering whenever \(s\) is strictly increasing.

\begin{proposition}\label{proposition_1}
    Assume \(F\) is a valid CDF with \(0<F(t)<1\) almost surely. Let \(r:[0,1]\to[0,1]\), and let \(s:[0,1]\to[0,1]\) be nondecreasing and right-continuous, with
    \[
    s(0)=0,
    \qquad
    \lim_{u\uparrow1}s(u)=1.
    \]
    Let \(F'\) be the recalibrated CDF in Equation~\eqref{eq:posttail_population_cdf}, let \(P'=1-F'(t)=r(P)\), and let \(Z'\) be the excess PIT induced by \(F'\). Then the operator has the following properties.
    
    \begin{enumerate}[leftmargin=1.5em]
        \item \(F'\) is a valid predictive CDF.
        \label{property:1}

        \item If \(r(P)=\Pr(E=1\mid P)\) almost surely, then \(P'\) satisfies occurrence calibration:
        \[
        \Pr(E=1\mid P')=P',
        \qquad
        \Pr(E=1)=\mathbb E[P'].
        \]
        \label{property:2}

        \item If \(s=H\), where \(H\) is the continuous conditional CDF of \(Z\mid E=1\), and \(P'>0\) almost surely on \(E=1\), then the recalibrated excess PIT satisfies
        \[
        Z'\mid E=1\sim \operatorname{Unif}(0,1).
        \]
        \label{property:3}

        \item If both population maps are used, then \(F'\) satisfies the combined fixed-threshold condition
        \[
        \Pr(Z'\le u,E=1)
        =
        u\,\mathbb E[P'],
        \qquad u\in[0,1].
        \]
        \label{property:4}

        \item The operator preserves the normalized body distribution whenever \(P'<1\):
        \[
        \frac{F'(y)}{F'(t)}
        =
        \frac{F(y)}{F(t)},
        \qquad y\le t.
        \]
        It also weakly preserves conditional tail ordering whenever \(s\) is nondecreasing, strictly preserves it when \(F\) and \(s\) are strictly increasing above the threshold, and reduces to the identity map when \(r(p)=p\) and \(s(u)=u\).
        \label{property:5}
    \end{enumerate}
\end{proposition}

\noindent
Proof sketch. Property~\ref{property:1} follows from monotonicity of the recombined CDF and the boundary conditions on \(r\), \(s\), and \(F\). Property~\ref{property:2} follows by the tower rule. Property~\ref{property:3} follows because the excess PIT of the recalibrated prediction is \(Z'=H(Z)\) on exceedances, and hence is uniform by the probability integral transform when \(H\) is continuous. Property~\ref{property:4} combines occurrence calibration with excess-PIT uniformity. The preservation statements in property~\ref{property:5} are direct algebraic consequences of the construction. Full details are given in Appendix~\ref{appdx:theory}. %

\section{PostTail: Estimating the Post-hoc Maps}
\label{sec:estimation}

The construction above is written in terms of population maps \(r\) and \(s\). In practice, our proposed method PostTail estimates these maps from a calibration set and then freezes them before final evaluation or deployment. The method is fit separately for each operational threshold and requires only evaluations of the issued predictive CDFs.

Let
\[
\mathcal D_{\mathrm{cal}}
=
\{(F_i,y_i)\}_{i=1}^{n}
\]
denote an observed calibration sample, typically a held-out validation set, where \(F_i\) is the predictive CDF issued for the realized outcome \(y_i\). The threshold \(t\) and all estimator settings are fixed before evaluating the method on the test sample. We assume that the calibration predictions are not used to train the base predictive model, that the final evaluation outcomes are not used to fit or select the post-hoc maps, and that calibration and future prediction--outcome pairs are exchangeable or identically distributed. The main presentation treats continuous predictive CDFs satisfying \(0<F_i(t)<1\).
For each calibration pair, compute
\[
q_i=F_i(t),\qquad
p_i=1-q_i,\qquad
e_i=\mathbf 1\{y_i>t\}.
\]
For calibration cases that exceed the threshold, define the excess-PIT values
\[
z_i=
\frac{F_i(y_i)-F_i(t)}{1-F_i(t)},
\qquad i:e_i=1.
\]
Thus occurrence calibration is a binary calibration problem based on all \(n\) pairs \((p_i,e_i)\), whereas severity calibration is a one-dimensional distributional calibration problem based on the \(m=\sum_i e_i\) excess-PIT values. The severity map is pooled over observed exceedances. It is not conditioned further on \(p_i\), covariates, or the full prediction.

\subsection{Occurrence-Map Estimation}

We estimate the occurrence map using beta calibration for binary probability predictions~\cite{kull2017beta}. The fitted map is
\begin{equation}
\widehat r(p)
=
\operatorname{logit}^{-1}
\{c+a\log p-b\log(1-p)\},
\quad a,b\ge0,
\label{eq:beta_occurrence_map}
\end{equation}
with parameters estimated by Bernoulli maximum likelihood on the calibration pairs \((p_i,e_i)\). The experiments use the monotone three-parameter beta family. The identity map is contained in this family because \(a\!=\!b\!=\!1\) and \(c\!=\!0\) give \(\widehat r(p)=p\). Beta calibration can therefore correct smooth systematic bias in predicted exceedance probabilities while using only a small number of parameters and without imposing the stepwise behavior of nonparametric isotonic regression.

For a new prediction \(F\), first compute the raw fitted probability
\[
\widehat p_{\mathrm{raw}}
=
\widehat r\{1-F(t)\}.
\]
The probability used in the recombined prediction is
\[
\widetilde p
=
\operatorname{clip}
\left(
\widehat p_{\mathrm{raw}},
\epsilon_p,
1-\epsilon_p
\right),
\]
where \(0<\epsilon_p<1/2\) is a fixed numerical safeguard that keeps the fitted probability away from zero and one. It does not create a missing body or conditional tail when the base prediction itself assigns zero mass to one side of \(t\).

\paragraph{Occurrence-map concentration assumption.}
Let \(r_\beta^\star\) denote a population Bernoulli-risk minimizer within the beta family, and write \(\widetilde r=C_{\epsilon_p}\circ\widehat r\) and \(r_{\beta,c}^\star=C_{\epsilon_p}\circ r_\beta^\star\) for the clipped fitted and population maps, where \(C_{\epsilon_p}(v)=\operatorname{clip}(v,\epsilon_p,1-\epsilon_p)\). To isolate the occurrence-estimation contribution in the fitted-map stability analysis below, we assume that, on a prespecified probability domain \(\mathcal P\), for every \(\delta_r\in(0,1)\) a deterministic error radius \(\varepsilon_r(n,\delta_r)\) satisfies, conditional on \(m\ge1\),
\[
\Pr\!\left\{
\left\|\widetilde r-r_{\beta,c}^\star\right\|_{\infty,\mathcal P}
\le
\varepsilon_r(n,\delta_r)
\;\middle|\;m\ge1
\right\}
\ge
1-\delta_r.
\]

We do not derive a beta-specific finite-sample form for \(\varepsilon_r\) here. The condition \(m\ge1\) only makes the severity map below well-defined. The bound may be uninformative for small \(m\). Appendix~\ref{appdx:posttail} defines the population target and discusses misspecification, clipping, conditioning, and regularity relevant to future refinements of the occurrence bound.

\subsection{Severity-Map Estimation}
Let \(m=\sum_i e_i\) be the number of calibration exceedances. When \(m>0\), we first apply a numerical endpoint convention to obtain endpoint-protected values \(\widetilde z_i\in(0,1)\) from the excess-PIT values above. The resulting right-continuous empirical CDF is
\begin{equation}
\widehat H_m(u)
=
\frac{1}{m}
\sum_{i:e_i=1}
\mathbf 1\{\widetilde z_i\le u\},
\qquad 0\le u\le1.
\label{eq:severity_ecdf}
\end{equation}
Endpoint protection ensures
\[
\widehat H_m(0)=0,
\qquad
\lim_{u\uparrow1}\widehat H_m(u)=1,
\]
as required by the validity conditions for the recombined CDF. We regularize the empirical map in Equation~\eqref{eq:severity_ecdf} toward the identity:
\begin{equation}
\widehat s_{\lambda_s}(u)
=
(1-\lambda_s)u
+
\lambda_s\widehat H_m(u),
\qquad 0\le\lambda_s\le1.
\label{eq:severity_regularization}
\end{equation}
Here \(\lambda_s=0\) leaves the conditional tail unchanged, whereas \(\lambda_s=1\) applies the full empirical correction. The convex combination remains nondecreasing and satisfies the required endpoint conditions.

The ECDF directly estimates \(\Pr(Z\le u\mid Y>t)\) without a parametric severity model. For \(\lambda_s\in(0,1)\), the map is affine between ECDF jumps rather than a pure step function, but its jumps still introduce atoms into the recalibrated tail. Consequently, CDF-based diagnostics and scores such as CRPS remain applicable, whereas an ordinary continuous-density log score is not appropriate without a mixed-distribution treatment or a smooth severity estimator.

\begin{lemma}\label{lem:severity-bound}
Suppose that, conditional on \(m\ge1\), the calibration excess-PIT values are independent draws from \(H\), and that endpoint protection is inactive almost surely. Then, for every \(\delta_s\in(0,1)\), with conditional probability at least \(1-\delta_s\),
\[
\left\|
\widehat s_{\lambda_s}-s_{\lambda_s}
\right\|_\infty
\le
\lambda_s
\sqrt{\frac{\log(2/\delta_s)}{2m}}.
\]

Consequently,
\[
\left\|
\widehat s_{\lambda_s}-H
\right\|_\infty
\le
(1-\lambda_s)\|I-H\|_\infty
+
\lambda_s
\sqrt{\frac{\log(2/\delta_s)}{2m}}.
\]
\end{lemma}

\begin{proof}
The proof applies the Dvoretzky-Kiefer-Wolfowitz inequality to \(\widehat H_m\) and uses the linear form of the identity regularization. 
\end{proof}

The full proof, together with the modification required when endpoint protection changes the theoretical excess-PIT values, is given in Appendix~\ref{appdx:theory}.

\subsection{Fitted PostTail Distribution}

For a new prediction \(F\) satisfying \(0<F(t)<1\), the fitted PostTail CDF is
\begin{equation}
\widehat F^{\mathrm{PT}}(y)=
\begin{cases}
(1-\widetilde p)\dfrac{F(y)}{F(t)},
& y\le t,\\[10pt]
1-\widetilde p
+\widetilde p\,\widehat s_{\lambda_s}\!\left(
\dfrac{F(y)-F(t)}{1-F(t)}
\right),
& y>t.
\end{cases}
\label{eq:fitted_posttail_cdf}
\end{equation}

Equation~\eqref{eq:fitted_posttail_cdf} is the population operator in Equation~\eqref{eq:posttail_population_cdf} with \(r\) and \(s\) replaced by their fitted counterparts and with occurrence-probability endpoint protection. The same fitted maps are reused for every test or deployment prediction at threshold \(t\).

For comparison, let \(F_{\lambda_s}^{\mathrm{PT},\star}\) denote the same recombination using the population maps \(r_{\beta,c}^\star\) and \(s_{\lambda_s}\), and let \(F_{\mathrm{oracle}}^{\mathrm{PT}}\) use the unrestricted population occurrence map \(r(p)=\Pr(E=1\mid P=p)\) and the unregularized severity map \(H\).

\begin{proposition}
\label{proposition_2}

Suppose \(m\ge1\), \(1-F(t)\in\mathcal P\), and the occurrence-map concentration assumption above and the conditions of Lemma~\ref{lem:severity-bound} hold. Then, for \(\delta_r,\delta_s\in(0,1)\) with \(\delta_r+\delta_s<1\), the following bounds hold with conditional probability at least \(1-\delta_r-\delta_s\).
\[
\left\|
\widehat F^{\mathrm{PT}}
-
F_{\lambda_s}^{\mathrm{PT},\star}
\right\|_\infty
\le
\varepsilon_r(n,\delta_r)
+
\lambda_s
\sqrt{\frac{\log(2/\delta_s)}{2m}}.
\]

Relative to the fully oracle recombination,
\[
\begin{aligned}
\left\|
\widehat F^{\mathrm{PT}}
-
F_{\mathrm{oracle}}^{\mathrm{PT}}
\right\|_\infty
\le{}&
\varepsilon_r(n,\delta_r)
+
\left\|r_{\beta,c}^\star-r\right\|_{\infty,\mathcal P}\\
&+
(1-\lambda_s)\|I-H\|_\infty
+
\lambda_s
\sqrt{\frac{\log(2/\delta_s)}{2m}}.
\end{aligned}
\]  
\end{proposition}

\begin{proof}
The recombination is Lipschitz, with constant one, in each map under the supremum norm, and then combines the two concentration events by a union bound.
\end{proof}

A full proof and the decomposition of the occurrence approximation term into beta-family misspecification and clipping effects are provided in Appendix~\ref{appdx:theory}.

\paragraph{}

PostTail requires at least one calibration exceedance and a base prediction with \(0<F(t)<1\). If no calibration exceedance is observed, both maps revert to the identity. If \(F(t)\in\{0,1\}\), the prediction lacks either a normalized body or a conditional tail and is handled by the prespecified boundary policy.
Proposition~\ref{proposition_2} quantifies how estimation of the two maps perturbs the recombined CDF, but it is a stability bound rather than a direct finite-sample calibration guarantee. Beta-family misspecification, fixed identity shrinkage, and clipping can leave nonvanishing approximation terms, and empirical calibration must therefore still be evaluated on held-out outcomes.
Further details on likelihood projection, regularization and alternative estimators for both maps, and the assumptions supporting the fitted-map results are provided in Appendix~\ref{appdx:posttail}. Full proofs of Lemma~\ref{lem:severity-bound} and Proposition~\ref{proposition_2} are provided in Appendix~\ref{appdx:theory}.

\section{Experiments and Results} \label{sec:experiments}

We conduct three simulation studies and a set of real-data experiments using UCI benchmark regression datasets. In this section we provide an overall setup and main findings from each study. %

\subsection{Simulation Studies}
We consider three complementary simulation studies that (i) verify the PostTail components and normalized-body preservation, (ii) demonstrate how pooled probability integral transform (PIT) calibration can conceal threshold-specific severity errors, and (iii) assess estimation and shrinkage when threshold exceedances are scarce. Full details and complete results are provided in Appendix~\ref{appdx:experiment_simulations}. Studies (i) and (ii) use severity-map weight \(\lambda_s=0.9\), selected empirically during method development and then held fixed.

\subsubsection{(i) Component-Wise Operator Verification}
At the fixed threshold \(t=0\), we construct analytical predictive CDFs with a \(\operatorname{Unif}(-1,0)\) subthreshold body, an exponential conditional upper tail, and heterogeneous exceedance probabilities. To examine component behavior and estimation effects, four scenarios isolate occurrence miscalibration, excess-severity miscalibration, both errors, and an already calibrated identity case. The corresponding PostTail components reduce their targeted held-out errors in the first three scenarios, while the recombined CDF preserves the normalized body. %
In the identity case, estimation can introduce sampling noise and slightly worsen mean held-out TMCB. %
The next study isolates a fixed-threshold error that remains invisible to the population global-PIT map. %

\subsubsection{(ii) Global PIT Calibration with Tail Miscalibration}

\begin{figure}
    \centering
    \includegraphics[width=0.95\linewidth]{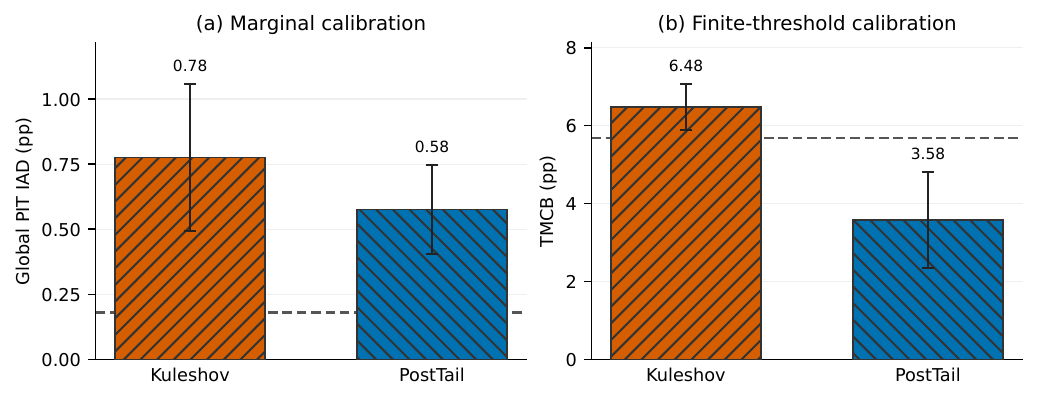}
    \caption{Comparison of Kuleshov's isotonic global recalibration and combined PostTail. Bars show the mean integrated absolute deviation of the global probability integral transform (global-PIT IAD) and the fixed-threshold calibration error (TMCB) across 10 repetitions. Values are multiplied by 100 and therefore represent average percentage-point deviations. Dashed horizontal lines mark the corresponding unmodified predictions, and error bars show $\pm 2$ standard errors across repetitions. Lower values are better.}
    \label{fig:scenario-e-global-vs-posttail}
\end{figure}

To examine whether marginal calibration can conceal operational-tail errors, we construct two equally likely prediction strata with correct exceedance rates but opposing conditional probability integral transform (PIT) distortions. These distortions cancel in the pooled PIT distribution, making the unmodified predictions globally PIT-calibrated in population, while their conditional excess-PIT values remain miscalibrated. Consequently, Kuleshov's global map has no population-level correction to learn and yields a mean TMCB of \(6.48\) percentage points, whereas combined PostTail directly targets the threshold components and reduces TMCB to \(3.58\) percentage points while retaining a low global-PIT IAD as shown in Fig.~\ref{fig:scenario-e-global-vs-posttail}. This controlled example illustrates global-PIT and fixed-threshold tail calibration targets are distinct and our method specifically addresses the latter.

\subsubsection{(iii) Finite-Sample Behavior under Tail Scarcity}
To assess finite-sample behavior, we vary the exceedance prevalence, the expected number of calibration exceedances \(n_t\), and the magnitude of occurrence and severity miscalibration. A severity-only study isolates estimation of the excess-PIT calibration map, whereas an end-to-end study jointly fits both PostTail components. With very few exceedances, unshrunk estimators can worsen calibration, including under identity and moderate-error settings. Count-based shrinkage substantially reduces this harm, while larger tail samples and stronger errors more consistently favor recalibration. %

\subsection{Real-Data Experiment}

\begin{table*}[t]
    \centering
    \setlength{\tabcolsep}{3pt}
    \resizebox{1.0\textwidth}{!}{
        \begin{tabular}{llcccccc}
        \toprule
        Training loss & Post-hoc method & Rel. CRPS $\downarrow$ & Rel. twCRPS $\downarrow$ & $|\Delta_{\mathrm{occ}}| \downarrow$ & Excess-IAD $\downarrow$ & TMCB $\downarrow$ & Global-IAD $\downarrow$ \\
        \midrule
        CRPS & None & $1.000$ & $1.000$ & $1.3$ & $14.1$ & $17.4$ & $4.8$ \\
        CRPS + $\lambda$ twCRPS ($\lambda$ selected per dataset) & None & $0.939$ & $0.924$ & $1.4$ & $13.0$ & $16.6$ & $5.7$ \\
        CRPS + $\gamma$ TMCB ($\gamma$ selected per dataset) & None & $1.056$ & $1.014$ & $1.3$ & $11.2$ & $15.0$ & $4.4$ \\
        \midrule
        CRPS & Global PIT & $0.995$ & $0.991$ & $1.3$ & $13.1$ & $16.5$ & $3.5$ \\
        CRPS + $\lambda$ twCRPS ($\lambda$ selected per dataset) & Global PIT & \underline{$0.931$} & \underline{$0.917$} & $1.2$ & $12.6$ & $16.6$ & \underline{$3.2$} \\
        CRPS + $\gamma$ TMCB ($\gamma$ selected per dataset) & Global PIT & $1.029$ & $1.025$ & $1.2$ & $12.1$ & $16.1$ & $\mathbf{3.0}$ \\
        \midrule
        CRPS & PostTail (ours) & $0.992$ & $0.967$ & \underline{$0.9$} & \underline{$10.7$} & $13.6$ & $4.6$ \\
        CRPS + $\lambda$ twCRPS ($\lambda$ selected per dataset) & PostTail (ours) & $\mathbf{0.930}$ & $\mathbf{0.888}$ & $\mathbf{0.8}$ & $\mathbf{10.3}$ & $\mathbf{12.9}$ & $5.4$ \\
        CRPS + $\gamma$ TMCB ($\gamma$ selected per dataset) & PostTail (ours) & $1.071$ & $1.039$ & $1.1$ & $\mathbf{10.3}$ & \underline{$13.4$} & $4.2$ \\
        \bottomrule
        \end{tabular}
    }
    \caption{Mean test performance over eight UCI datasets and five seeds. CRPS and twCRPS are relative to the unmodified CRPS-trained model. A value below one is better. Calibration metrics are \(\times 100\). Hyperparameters are selected using validation data. Lower is better.}
    \label{tab:uci-tail-main-summary}
    \vspace{-0.65em}
\end{table*}

\subsubsection{Setup.}
We evaluate whether PostTail improves calibration at one prespecified operational threshold, and whether post-hoc recalibration provides a useful alternative or complement to tail-aware model training. The study uses eight tabular regression benchmarks obtained from the UCI Machine Learning Repository~\cite{uci2023}: Boston Housing, Concrete, Energy, Kin8nm, Naval, Power, Protein, and Yacht. These established benchmarks span a wide range of sample sizes and realized tail support, allowing the same protocol to be evaluated under heterogeneous data regimes.
For each dataset and five random seeds, we use approximately \(72\%\), \(18\%\), and \(10\%\) of the observations for training, validation/calibration, and testing, respectively. 
Features and responses are standardized using the training split only, and \(t\) is fixed at the empirical \(0.9\) quantile of that split. The test split is used only for final reporting.

\subsubsection{Baselines.}
We use the same two-layer, width-64 MLP under three training objectives.
\[
\operatorname{CRPS},\quad
\operatorname{CRPS}+\lambda \operatorname{twCRPS}_t,\quad
\operatorname{CRPS}+\gamma \operatorname{TMCB}_t.
\]
The CRPS+twCRPS objective emphasizes upper-tail predictive performance, whereas CRPS+TMCB encourages fixed-threshold calibration following~\cite{wessel2026enforcing}. For each dataset and seed, we first train one CRPS model, then initialize from it and fine-tune nine CRPS+twCRPS candidates and five CRPS+TMCB candidates. One \(\lambda\) and one \(\gamma\) are selected per dataset by grid search on the validation set using twCRPS and TMCB, respectively.

For each selected trained model, we compare the unmodified predictive CDF, global-PIT recalibration following~\cite{kuleshov2018accurate}, and PostTail. Both recalibrators are fit on the calibration split. PostTail uses occurrence weight \(\lambda_r=1\) and severity weight \(\lambda_s=0.9\). These recalibration weights are distinct from the selected training-loss coefficients \(\lambda\) and \(\gamma\) and the \(0.9\)-quantile level defining the threshold. TMCB is the primary metric because it directly summarizes the combined fixed-threshold occurrence and severity target. Occurrence gap and Excess-IAD diagnose its two components, Global-IAD diagnoses pooled PIT calibration, and CRPS and twCRPS check full-distribution and upper-tail predictive performance. Table~\ref{tab:uci-tail-main-summary} reports the results across all datasets. %

\subsubsection{Results.}
Compared with the unmodified predictions, PostTail lowers mean TMCB in each of the three training families in Table~\ref{tab:uci-tail-main-summary}. Both mean proper scores also improve for the CRPS and CRPS+twCRPS families, but the CRPS+TMCB family exhibits a trade-off: mean TMCB decreases from \(15.0\) to \(13.4\) percentage points, while relative CRPS increases from \(1.056\) to \(1.071\) and relative twCRPS from \(1.014\) to \(1.039\). Thus, the aggregate fixed-threshold calibration improvements do not imply improvements in both proper scores for every training family. Applied to the CRPS-trained model, it reduces mean TMCB from \(17.4\) to \(13.6\) percentage points. The selected CRPS+twCRPS model followed by PostTail yields the lowest TMCB in Table~\ref{tab:uci-tail-main-summary}, \(12.9\), together with relative CRPS \(0.930\) and relative twCRPS \(0.888\). The component diagnostics show the same general pattern. Across the \(120\) paired dataset--seed--training-family comparisons, PostTail has lower TMCB than global PIT in \(66.7\%\) of cases, while it has lower Global-IAD in \(38.3\%\), as reported in Table~\ref{tab:uci-tail-main-posttail-vs-tail-aware-win-rates}. 
These win counts summarize paired comparisons across five seed runs per dataset. 
Its occurrence-gap and Excess-IAD win rates are \(63.3\%\) and \(69.2\%\), respectively. Relative to the unmodified predictions, its mean Global-IAD decreases from \(5.0\) to \(4.7\) percentage points across the selected training families, while mean TMCB decreases from \(16.3\) to \(13.3\). Global PIT has lower mean Global-IAD than PostTail in each training family, as expected from the methods' different targets: PostTail addresses occurrence and excess-severity calibration at the fixed threshold, rather than global-PIT calibration.
PostTail directly concentrates its calibration capacity on exceedance probabilities and excess-PIT values at \(t\), unlike global PIT's single map over the full predictive CDF. The observed pattern of more frequent TMCB wins and less frequent Global-IAD wins is consistent with these different targets. These results illustrate empirically that pooled PIT calibration alone does not guarantee fixed-threshold tail calibration.

PostTail is also competitive when using the CRPS-trained model as a common base. It has lower test TMCB than the selected training-only CRPS+twCRPS model in \(27/40\) paired dataset--seed comparisons (\(67.5\%\)) and than the selected training-only CRPS+TMCB model in \(23/40\) comparisons (\(57.5\%\)). These comparisons are reported in Table~\ref{tab:uci-tail-main-posttail-vs-tail-aware-win-rates}. They support PostTail as a post-hoc counterpart to, rather than a replacement for, these training-time alternatives.

\begin{table}[t]
\centering
\resizebox{1.0\linewidth}{!}{
\begin{tabular}{lccc}
\toprule
Metric & Global PIT & CRPS+$\lambda$twCRPS & CRPS+$\gamma$TMCB \\
\midrule
CRPS & 58/120 (48.3\%) & 20/40 (50.0\%) & 9/40 (22.5\%) \\
twCRPS & 74/120 (61.7\%) & 18/40 (45.0\%) & 19/40 (47.5\%) \\
$|\Delta_{\mathrm{occ}}|$ & 76/120 (63.3\%) & 28/40 (70.0\%) & 26/40 (65.0\%) \\
Excess-IAD & 83/120 (69.2\%) & 26/40 (65.0\%) & 23/40 (57.5\%) \\
TMCB & 80/120 (66.7\%) & 27/40 (67.5\%) & 23/40 (57.5\%) \\
Global-IAD & 46/120 (38.3\%) & 24/40 (60.0\%) & 16/40 (40.0\%) \\
\bottomrule
\end{tabular}
}
\caption{Test-set win rate for PostTail against global-PIT recalibration and models trained with tail-aware objectives. Entries report count/total (percentage). Ties are non-wins.}
\label{tab:uci-tail-main-posttail-vs-tail-aware-win-rates}
\vspace{-0.65em}
\end{table}

\section{Conclusion} \label{sec:conclusion}

We introduced PostTail, a model-agnostic method that uses held-out data to recalibrate exceedance probabilities and excess-PIT values at one prespecified threshold, recombining them into a valid predictive CDF while preserving normalized subthreshold shape. Population maps attain the stated marginal fixed-threshold calibration targets, whereas the fitted-map theory provides stability bounds rather than exact finite-sample guarantees. The empirical studies show  fixed-threshold improvements, with proper-score trade-offs and no uniform advantage in global-PIT calibration, particularly when exceedances are scarce. PostTail therefore provides a focused post-hoc alternative when retraining is impractical. The real-data evaluation uses one base-model architecture, one fixed threshold, and a limited set of post-hoc baselines. Occurrence-reliability diagnostics on real data, broader model and baseline comparisons, and evaluation at application-defined thresholds remain future work, alongside improved estimation when tail samples are scarce.

\section*{Acknowledgments}
This work was supported by the Estonian Research Council grant PRG1604, by the Estonian Academy of Sciences research professorship in AI, and by the Estonian Center of Excellence in Artificial Intelligence (EXAI), funded by the Estonian Ministry of Education and Research grant TK213.

\bibliography{references}

\onecolumn
\appendix
\section*{Appendix}

\section{Proofs and Theoretical Details}\label{appdx:theory}
\subsection{Proof of Proposition~\ref{proposition_1}}
\label{appdx:proof_proposition_1}
\begin{proof}
    We verify the stated properties in order.\\

    \noindent
    \textbf{Property~\ref{property:1}. Validity of the recalibrated CDF}\\
    Fix a realized prediction \(F\), write \(q=F(t)\in(0,1)\), and write \(P'=r(1-q)\in[0,1]\). Define
    \[
    U(y)=\frac{F(y)-q}{1-q},
    \qquad y>t.
    \]
    For \(y\le t\), since \(0\le F(y)\le q\),
    \[
    0\le
    (1-P')\frac{F(y)}{q}
    \le
    1-P'.
    \]
    For \(y>t\), since \(q\le F(y)\le1\), \(U(y)\in[0,1]\). Since \(s\) maps \([0,1]\) into \([0,1]\),
    \[
    1-P'
    \le
    1-P'+P's(U(y))
    \le
    1.
    \]
    Thus \(F'\) takes values in \([0,1]\).
    The lower piece is nondecreasing because it is a nonnegative scalar multiple of the CDF \(F\). The upper piece is nondecreasing because \(U(y)\) is nondecreasing and \(s\) is nondecreasing. Across the threshold, for \(y_1\le t<y_2\),
    \[
    F'(y_1)\le1-P'\le F'(y_2).
    \]
    Therefore \(F'\) is nondecreasing on the real line.
    The lower formula gives
    \[
    F'(t)=(1-P')\frac{F(t)}{F(t)}=1-P'.
    \]
    For \(y\downarrow t\) with \(y>t\), right-continuity of \(F\) gives \(F(y)\to F(t)\), so \(U(y)\to0\). Since \(s\) is right-continuous at zero and \(s(0)=0\),
    \[
    \lim_{y\downarrow t,\,y>t}F'(y)=1-P'=F'(t).
    \]
    Right-continuity away from \(t\) follows from right-continuity of \(F\) and \(s\). Hence \(F'\) is right-continuous everywhere.
    Finally,
    \[
    \lim_{y\to-\infty}F'(y)=0.
    \]
    As \(y\to\infty\), \(U(y)\to1\). By the endpoint condition in Proposition~\ref{proposition_1},
    \[
    \lim_{y\to\infty}F'(y)
    =
    1-P'+P'\lim_{u\uparrow1}s(u)
    =1.
    \]
    Thus \(F'\) is nondecreasing, right-continuous, and has limits zero and one, so it is a valid CDF. \\

    \noindent
    \textbf{Property 2. Population occurrence calibration}\\
    If \(r(P)=\Pr(E=1\mid P)\) almost surely, then by the tower property,
    \[
    \mathbb E[E\mid P']
    =
    \mathbb E\{\mathbb E[E\mid P]\mid P'\}
    =
    \mathbb E\{r(P)\mid r(P)\}
    =P'.
    \]
    Taking expectations gives \(\Pr(E=1)=\mathbb E[P']\). This proves property 2. \\

    \noindent
    \textbf{Property 3. Population excess-PIT calibration}\\
    Since \(F'(t)=1-P'\), for \(y>t\) and \(P'>0\),
    \[
    \frac{F'(y)-F'(t)}{1-F'(t)}
    =
    H(F_t^+(y)).
    \]
    Thus, on exceedances with \(P'>0\), \(Z'=H(Z)\). If \(H\) is continuous, the probability integral transform gives
    \[
    H(Z)\mid E=1\sim\operatorname{Unif}(0,1).
    \]
    Under the population occurrence map in property 2, the positivity condition holds automatically up to null events, because
    \[
    \Pr(E=1,P'=0)=\mathbb E[\mathbf 1\{P'=0\}P']=0.
    \]
    This proves property 3. \\

    \noindent
    \textbf{Property 4. Combined fixed-threshold calibration}\\
    By property 3,
    \[
    \Pr(Z'\le u\mid E=1)=u.
    \]
    By property 2, \(\Pr(E=1)=\mathbb E[P']\). Therefore
    \[
    \Pr(Z'\le u,E=1)
    =u\Pr(E=1)
    =u\mathbb E[P'].
    \]
    This proves property 4. \\

    \noindent
    \textbf{Property 5. Preservation properties}\\
    Since \(F'(t)=1-P'\), for \(y\le t\) and \(P'<1\),
    \[
    \frac{F'(y)}{F'(t)}
    =
    \frac{(1-P')F(y)/F(t)}{1-P'}
    =
    \frac{F(y)}{F(t)}.
    \]
    Thus, the operator changes the total body mass but not the normalized body shape. Tail ordering follows because the recalibrated conditional tail CDF is \(s\circ F_t^+\). Finally, if \(r(p)=p\) and \(s(u)=u\), then \(1-P'=F(t)\), so both pieces of Equation~\eqref{eq:posttail_population_cdf} reduce algebraically to \(F(y)\). \\
    This completes the proof of Proposition~\ref{proposition_1}. 
\end{proof}

\subsection{Boundary Conditions and Scope}
\label{appdx:boundary_conditions}

The proposition assumes \(0\le r(p)\le1\), a nondecreasing right-continuous severity map \(s\), and \(s(0)=0\), \(\lim_{u\uparrow1}s(u)=1\). These conditions ensure that the recombined prediction is a valid CDF.
The recombination in Equation~\eqref{eq:posttail_population_cdf} is well defined when \(0<F(t)<1\). If \(P=0\), the conditional upper-tail distribution is undefined. If \(F(t)=0\), the body normalization is undefined. Such boundary cases are handled by a prespecified convention, such as leaving the affected prediction unchanged or applying a small probability floor. In sample-based predictions, the required CDF values can be estimated using empirical distribution functions.

For an empirical severity map \(s\), the recalibrated CDF may have jumps in the tail. This is still a valid predictive CDF, but smoother estimators may be preferable when density evaluation, log scores, or smooth quantiles are required.
If the occurrence map is constrained to be monotone but the true conditional probability \(p\mapsto\Pr(E=1\mid P=p)\) is not monotone, an isotonic population estimator targets a monotone projection rather than the unrestricted population map in Equation~\eqref{eq:occurrence_map}. In that case, the exact occurrence-calibration statement in property~\ref{property:2} of Proposition~\ref{proposition_1} applies only to the projected map under additional assumptions.

If \(H\) is discontinuous, the nonrandomized transform \(H(Z)\) need not be uniform. The standard randomized PIT restores exact uniformity. This matters for empirical CDF severity maps and for predictions or outcomes with atoms.

\subsection{Randomized Excess PIT for Discontinuous CDF \(H\)}
\label{appdx:randomized_excess_PIT}
If \(H\) is discontinuous, the ordinary transform \(H(Z)\) is generally not uniform. Let \(V\sim\operatorname{Unif}(0,1)\) be independent of \((F,Y)\), and define
\[
Z'_{\mathrm{rand}}
=
H(Z^-)+V\{H(Z)-H(Z^-)\},
\qquad
H(z^-)=\lim_{v\uparrow z}H(v).
\]
The randomized probability integral transform gives
\[
Z'_{\mathrm{rand}}\mid E=1\sim\operatorname{Unif}(0,1).
\]
Therefore, the exact excess-PIT conclusion remains valid for discontinuous conditional excess-PIT distributions in the randomized-PIT sense.

\subsection{Proof of Lemma~\ref{lem:severity-bound}}

Suppose that, conditional on \(m\ge1\), the calibration excess-PIT values are independent draws from the CDF \(H\). Let \(\widehat H_m\) be their empirical CDF, let
\[
\widehat s_{\lambda_s}
=
(1-\lambda_s)I+\lambda_s\widehat H_m,
\qquad
s_{\lambda_s}
=
(1-\lambda_s)I+\lambda_s H,
\]
where \(I(u)=u\). Then, for every \(\delta\in(0,1)\), with conditional probability at least \(1-\delta\),
\[
\left\|
\widehat s_{\lambda_s}-s_{\lambda_s}
\right\|_\infty
\le
\lambda_s
\sqrt{
\frac{\log(2/\delta)}{2m}
}.
\]
Consequently,
\[
\left\|
\widehat s_{\lambda_s}-H
\right\|_\infty
\le
(1-\lambda_s)\left\|I-H\right\|_\infty
+
\lambda_s
\sqrt{
\frac{\log(2/\delta)}{2m}
}.
\]

\begin{proof}
    The Dvoretzky--Kiefer--Wolfowitz inequality, with the sharp constant established by Massart, gives
    \[
    \Pr\!\left(
    \left\|\widehat H_m-H\right\|_\infty>\varepsilon
    \;\middle|\;m
    \right)
    \le
    2\exp(-2m\varepsilon^2)
    \]
    for every \(\varepsilon>0\)~\cite{dvoretzky1956asymptotic,massart1990tight}. Since
    \[
    \widehat s_{\lambda_s}-s_{\lambda_s}
    =
    \lambda_s(\widehat H_m-H),
    \]
    setting \(\varepsilon=\sqrt{\log(2/\delta)/(2m)}\) proves the first bound. The second follows from
    \[
    \widehat s_{\lambda_s}-H
    =
    (1-\lambda_s)(I-H)
    +
    \lambda_s(\widehat H_m-H)
    \]
    and the triangle inequality.
\end{proof}

\subsection{Endpoint-Protected Excess-PIT}

The underlying concentration result is classical. Lemma~\ref{lem:severity-bound} is its direct application to the regularized PostTail severity map. The second bound separates the deterministic bias introduced by identity regularization from the sampling error controlled by the number of calibration exceedances.\\
The lemma in the main text assumes that endpoint protection is inactive almost surely. More generally, let \(T:[0,1]\to[0,1]\) be a fixed endpoint-protection map, let \(\widetilde Z=T(Z)\), and let
\[
H_T(u)
=
\Pr\{\widetilde Z\le u\mid E=1\}.
\]
The endpoint-protected excess-PIT values are independent draws from \(H_T\), so the same DKW argument gives
\[
\left\|
\widehat s_{\lambda_s}
-
\{(1-\lambda_s)I+\lambda_s H_T\}
\right\|_\infty
\le
\lambda_s
\sqrt{\frac{\log(2/\delta_s)}{2m}}
\]
with conditional probability at least \(1-\delta_s\). Relative to the original population severity map \(H\),
\[
\begin{aligned}
\left\|
\widehat s_{\lambda_s}-H
\right\|_\infty
\le{}&
(1-\lambda_s)\|I-H\|_\infty
+
\lambda_s\|H_T-H\|_\infty\\
&+
\lambda_s
\sqrt{\frac{\log(2/\delta_s)}{2m}}.
\end{aligned}
\]
Thus endpoint protection changes the population target only when it changes theoretical excess-PIT values with positive probability. Under the continuous interior-PIT assumption used in the main statement, \(T(Z)=Z\) almost surely and the additional term is zero.

\subsection{Proof of Proposition 2}
\label{appdx:proof_proposition_2}

\begin{proof}
    For a fixed base prediction \(F\), write \(q=F(t)\in(0,1)\) and define the recombination operator
    \[
    \mathcal T_F(p,s)(y)
    =
    \begin{cases}
    (1-p)\dfrac{F(y)}{q},
    &y\le t,\\[9pt]
    1-p+p\,s\!\left(
    \dfrac{F(y)-q}{1-q}
    \right),
    &y>t.
    \end{cases}
    \]
    Consider two occurrence--severity pairs \((p_1,s_1)\) and \((p_0,s_0)\). For \(y\le t\),
    \[
    \left|
    \mathcal T_F(p_1,s_1)(y)
    -
    \mathcal T_F(p_0,s_0)(y)
    \right|
    =
    |p_1-p_0|\frac{F(y)}{q}
    \le
    |p_1-p_0|.
    \]
    For \(y>t\), let \(u=\{F(y)-q\}/(1-q)\). Direct subtraction gives
    \[
    \begin{aligned}
    \mathcal T_F(p_1,s_1)(y)
    -
    \mathcal T_F(p_0,s_0)(y)
    ={}&
    (p_0-p_1)\{1-s_0(u)\}\\
    &+
    p_1\{s_1(u)-s_0(u)\}.
    \end{aligned}
    \]
    Since \(p_1\in[0,1]\) and \(s_0(u)\in[0,1]\),
    \begin{equation}
    \left\|
    \mathcal T_F(p_1,s_1)
    -
    \mathcal T_F(p_0,s_0)
    \right\|_\infty
    \le
    |p_1-p_0|
    +
    \|s_1-s_0\|_\infty.
    \label{eq:recombination_lipschitz}
    \end{equation}
    Apply Equation~\eqref{eq:recombination_lipschitz} with
    \[
    p_1
    =
    \widetilde r\{1-F(t)\},
    \qquad
    p_0
    =
    r_{\beta,c}^\star\{1-F(t)\},
    \qquad
    s_1
    =
    \widehat s_{\lambda_s},
    \qquad
    s_0
    =
    s_{\lambda_s}.
    \]
    On the occurrence concentration event and the event in Lemma~\ref{lem:severity-bound},
    \[
    \left\|
    \widehat F^{\mathrm{PT}}
    -
    F_{\lambda_s}^{\mathrm{PT},\star}
    \right\|_\infty
    \le
    \varepsilon_r(n,\delta_r)
    +
    \lambda_s
    \sqrt{\frac{\log(2/\delta_s)}{2m}}.
    \]
    The union bound gives simultaneous probability at least \(1-\delta_r-\delta_s\). Independence between the two fitted maps is not required.\\
    For the comparison with the fully oracle recombination, apply Equation~\eqref{eq:recombination_lipschitz} once more to obtain
    \[
    \left\|
    F_{\lambda_s}^{\mathrm{PT},\star}
    -
    F_{\mathrm{oracle}}^{\mathrm{PT}}
    \right\|_\infty
    \le
    \left\|
    r_{\beta,c}^\star-r
    \right\|_{\infty,\mathcal P}
    +
    \left\|
    s_{\lambda_s}-H
    \right\|_\infty.
    \]
    Because
    \[
    \left\|
    s_{\lambda_s}-H
    \right\|_\infty
    =
    (1-\lambda_s)\|I-H\|_\infty,
    \]
    the second inequality in Proposition 2 follows by the triangle inequality. If endpoint protection is active at the population level, the additional term \(\lambda_s\|H_T-H\|_\infty\) derived above must also be included. This completes the proof.
\end{proof}

\subsection{Occurrence Approximation Term}

Let \(r_\beta^\star\) be the population Bernoulli-risk minimizer within the beta family, let \(C_{\epsilon_p}(v)=\operatorname{clip}(v,\epsilon_p,1-\epsilon_p)\), and write \(r_{\beta,c}^\star=C_{\epsilon_p}\circ r_\beta^\star\). When comparing Proposition~\ref{proposition_2} with the unrestricted population occurrence map \(r(p)=\Pr(E=1\mid P=p)\),
\[
\left\|
r_{\beta,c}^\star-r
\right\|_{\infty,\mathcal P}
\le
\left\|
r_\beta^\star-r
\right\|_{\infty,\mathcal P}
+
\epsilon_p.
\]
The first term is the beta-family approximation error and vanishes when the occurrence map is correctly specified and the population minimizer is identified on \(\mathcal P\). The clipping contribution is zero whenever \(r_\beta^\star(p)\in[\epsilon_p,1-\epsilon_p]\) for every \(p\in\mathcal P\).

\section{PostTail Estimation and Evaluation Details}
\label{appdx:posttail}

Appendix~\ref{appdx:posttail} provides supporting details for the fitted PostTail estimators and complements the formal proofs in Appendix~\ref{appdx:theory}. It first identifies the population target of beta occurrence estimation. It then states the occurrence-map concentration condition used in Proposition~2, with the estimator's finite-sample accuracy represented by the radius $\varepsilon_r(n,\delta_r)$. The accompanying discussion outlines the regularity conditions needed for a beta-specific bound and separates estimation error from approximation error caused by beta-family misspecification and clipping. The remainder of Appendix~\ref{appdx:posttail} covers regularization and alternative estimators for the occurrence and severity maps, the population effect of severity shrinkage, the scope of separate threshold-wise fitting, and the evaluation metrics used in the experiments.

\subsection{Population Target of Beta Occurrence Estimation}

Write \(\theta=(a,b,c)\) and let \(r_\theta\) denote the beta-calibration map in Equation~\eqref{eq:beta_occurrence_map}. Its estimator solves
\[
\widehat\theta
\in
\arg\min_{a,b\ge0,\;c\in\mathbb R}
\widehat L_n(\theta),
\]
where
\[
\widehat L_n(\theta)
=
-
\frac{1}{n}
\sum_{i=1}^{n}
\left[
e_i\log r_\theta(p_i)
+
(1-e_i)\log\{1-r_\theta(p_i)\}
\right].
\]
Under the usual conditions for consistency of constrained \(M\)-estimators, \(\widehat\theta\) converges to a minimizer of the population Bernoulli risk
\[
\theta^\star
\in
\arg\min_{a,b\ge0,\;c\in\mathbb R}
\mathbb E
\left[
-E\log r_\theta(P)
-(1-E)\log\{1-r_\theta(P)\}
\right].
\]
This characterization identifies the population quantity approached by the estimated occurrence map. For the prespecified evaluation domain \(\mathcal P\subset(0,1)\) used below, if \(p\mapsto\Pr(E=1\mid P=p)\) belongs to the beta family and the population minimizer is identified on \(\mathcal P\), then \(r_{\theta^\star}(p)=\Pr(E=1\mid P=p)\) for \(p\in\mathcal P\). Otherwise, \(r_{\theta^\star}\) is its likelihood projection onto that family. The distinction clarifies why the population guarantee in Proposition~\ref{proposition_1} does not automatically transfer to a misspecified or unidentified occurrence estimator.

\subsection{Occurrence-Map Concentration Assumption}

On the prespecified domain \(\mathcal P\), write \(r_\beta^\star=r_{\theta^\star}\). The concentration assumption used in Proposition~\ref{proposition_2} is
\[
\Pr\!\left\{
\left\|
\widetilde r-r_{\beta,c}^\star
\right\|_{\infty,\mathcal P}
\le
\varepsilon_r(n,\delta_r)
\;\middle|\;m\ge1
\right\}
\ge
1-\delta_r,
\]
where
\[
\widetilde r
=
C_{\epsilon_p}\circ\widehat r,
\qquad
r_{\beta,c}^\star
=
C_{\epsilon_p}\circ r_\beta^\star,
\qquad
C_{\epsilon_p}(v)
=
\operatorname{clip}(v,\epsilon_p,1-\epsilon_p).
\]
This is a modular assumption on the occurrence estimator, rather than a finite-sample theorem proved here. A beta-specific uniform bound would require, among other conditions, a compact calibration domain bounded away from zero and one, identification of the population parameter, sufficient curvature of the Bernoulli risk, and a nondegenerate distribution of the predicted probabilities. Under these and the usual conditions for constrained \(M\)-estimation, parameter consistency and continuity of the beta map yield uniform asymptotic convergence on \(\mathcal P\). If \(\Pr(E=1)>0\), then \(\Pr(m\ge1)\to1\), so the same asymptotic conclusion holds under the conditional law used in Proposition~\ref{proposition_2}. These asymptotic statements do not by themselves provide an explicit finite-\(n\) radius.

The clipping map \(C_{\epsilon_p}\) is one-Lipschitz, and hence
\[
\left\|
\widetilde r-r_{\beta,c}^\star
\right\|_{\infty,\mathcal P}
\le
\left\|
\widehat r-r_\beta^\star
\right\|_{\infty,\mathcal P}.
\]
Thus any uniform concentration result for the unprotected beta estimator transfers directly to the protected maps. Because the severity result is conditional on observing at least one exceedance, the probability statements in Proposition~\ref{proposition_2} are understood under the same calibration-sample law conditional on \(m\ge1\). Equivalently, the occurrence-map concentration assumption must hold under that law. Appendix~\ref{appdx:theory} decomposes the remaining approximation relative to the unrestricted population occurrence map into beta-family approximation and clipping terms.

\subsection{Regularization of the Two Maps}
Identity regularization can be applied to either estimated map. For the occurrence component, define
\[
\widehat r_{\lambda_r}(p)
=
(1-\lambda_r)p
+
\lambda_r\widehat r(p),
\qquad
\lambda_r\in[0,1].
\]
Equation~\eqref{eq:severity_regularization} applies the analogous construction to the severity map. Both forms preserve monotonicity and interpolate between no recalibration and the fully fitted map. The need for explicit shrinkage nevertheless differs between estimators. Beta calibration already contains the identity map and uses a low-dimensional parametric family. We therefore use the full fitted beta map, \(\lambda_r=1\), in the primary method. This identity inclusion prevents a structural inability to represent an already calibrated prediction, but it does not eliminate finite-sample estimation error. Occurrence shrinkage may still be useful with very small or highly imbalanced calibration samples.

Isotonic occurrence calibration is substantially more flexible and can produce many data-dependent constant blocks. Although its monotone function class can represent the identity, the empirical fit may deviate sharply from it in small samples. Explicit shrinkage toward \(p\), block smoothing, or another regularized isotonic estimator is therefore more consequential for isotonic calibration than for the three-parameter beta map. For the ECDF severity estimator, regularization is motivated directly by the effective sample size \(m\), which can be much smaller than \(n\).

\subsection{Estimator Choices and Alternatives}

Possible alternatives to the primary beta estimator include increasing isotonic regression and a one-parameter logit-offset map. The logit-offset map can match the aggregate exceedance rate while preserving probability ordering, but it is not flexible enough to estimate a general reliability curve \(p\mapsto\Pr(E=1\mid P=p)\). 

Alternatives to the empirical severity map include monotone linear or spline estimators with fixed endpoints, beta-CDF families, kernel-smoothed distribution estimators, and Bayesian distribution estimators. These alternatives trade the ECDF's directness and distribution-free behavior for smoothness, lower variance, or stronger modeling assumptions. Discrete predictions or structural ties additionally require a randomized distributional transform.
The empirical severity map can also be viewed as the conditional-tail version of the regression recalibration method of~\cite{kuleshov2018accurate}. For an exceedance, let \(Z=G(Y)\) denote the PIT under the base conditional-tail CDF \(G\). The Kuleshov recalibration target, restricted to exceedances, is
\[
\Pr\!\left\{Y\le G^{-1}(u)\mid Y>t\right\}
=
\Pr(Z\le u\mid Y>t)
=
H(u),
\]
which is precisely the population severity map. The ECDF of the excess-PIT values is therefore the canonical nonparametric plug-in estimator of this PIT-calibration map. \citet{kuleshov2018accurate} recommend isotonic regression on PIT values and their corresponding empirical cumulative probabilities. This estimator and the ECDF target the same population function, although their finite-sample representations can differ through interpolation, endpoint treatment, or smoothing conventions. The endpoint-protected and identity-regularized ECDF used by PostTail is thus a regularized PIT calibrator fitted only to threshold exceedances.

\subsection{Population Effect of Severity Regularization}

The population counterpart of Equation~\eqref{eq:severity_regularization} is
\[
s_{\lambda_s}(u)
=
(1-\lambda_s)u+\lambda_s H(u).
\]
For a fixed \(\lambda_s<1\), the estimator targets \(s_{\lambda_s}\), not the oracle map \(H\). Exact population excess-PIT uniformization is therefore recovered only when \(\lambda_s=1\), when \(\lambda_s\to1\) as calibration information increases, or when \(H(u)=u\). A fixed value below one deliberately retains part of the original conditional-tail distribution in exchange for reduced estimation variability.

\subsection{Separate Estimation Across Thresholds}

PostTail estimates a different occurrence--severity map pair for each prespecified threshold. These separately fitted operators need not agree on their implied probabilities across thresholds. Consequently, the single-threshold construction does not by itself provide simultaneous or coherent multi-threshold calibration. Developing a joint estimator that enforces such coherence is left for future work.

\subsection{Evaluation Metrics}
\label{appdx:experiment_metrics}

Let \(F_i\) denote the predictive CDF on test case \(i\), let \(p_i=1-F_i(t)\), and let \(e_i=\mathbf1\{y_i>t\}\). When \(e_i=1\), let \(z_i=\{F_i(y_i)-F_i(t)\}/\{1-F_i(t)\}\).

\subsubsection{Occurrence Gap.} For occurrence, let the signed marginal occurrence gap be
\[
\Delta_{\mathrm{occ}}
=
\frac{1}{n_{\mathrm{test}}}\sum_i e_i
-
\frac{1}{n_{\mathrm{test}}}\sum_i p_i,
\]
and report \(\left|\Delta_{\mathrm{occ}}\right|\). This is the absolute difference between the empirical test exceedance rate and the mean predicted exceedance probability. It measures marginal occurrence agreement, allowing errors at different issued probabilities to cancel. Occurrence reliability instead requires assessing exceedance frequencies across issued probabilities, for example through held-out reliability diagrams and occurrence ECE computed using all test cases, including non-exceedances. 
Before taking the absolute value, \(\Delta_{\mathrm{occ}}>0\) indicates underprediction of exceedances and \(\Delta_{\mathrm{occ}}<0\) overprediction. %
\subsubsection{Excess Integral Absolute Deviation.} severity, or excess, calibration is evaluated among observed test exceedances as follows:
\[
\operatorname{Severity/Excess-IAD}_t
=
\int_0^1
\left|\widehat H_{\mathrm{test}}(u)-u\right|\,du.
\]

\subsubsection{Global Integral Absolute Deviation.} Similar to Excess-IAD, the global integral absolute deviation (Global-IAD) assesses calibration of the issued predictive CDFs over the full distribution rather than only the excess region. Let \(\widehat F_{\mathrm{test}}\) be the empirical CDF of the PIT values \(F_i(y_i)\) over all test cases. Then
\[
\operatorname{Global-IAD}
=
\int_0^1
\left|\widehat F_{\mathrm{test}}(u)-u\right|\,du.
\]

\subsubsection{Tail Measure of Miscalibration.} The combined occurrence--severity miscalibration is assessed by the tail measure of miscalibration (TMCB)~\cite{wessel2026enforcing}. 
Let \(\widehat R_t\) be the empirical combined fixed-threshold calibration curve based on the test pairs \((e_i,z_i)\), with \(z_i\) defined only for \(e_i=1\).
\[
\widehat R_t(u)
=
\frac{
\sum_i e_i\mathbf1\{z_i\le u\}
}{
\sum_i p_i
},
\qquad
\operatorname{TMCB}_t
=
\int_0^1|\widehat R_t(u)-u|\,du.
\]
TMCB summarizes the marginal fixed-threshold target, combining occurrence matching and excess-PIT uniformity pooled over prediction instances. Even a zero population TMCB does not imply occurrence reliability across issued probabilities. Finite-sample reliability diagrams and binned occurrence ECE provide evidence about this stronger property at the resolution supported by the data, rather than establishing exact population reliability. We therefore report it together with \(\left|\Delta_{\mathrm{occ}}\right|\) and \(\operatorname{Excess\mbox{-}IAD}_t\), which localize occurrence-mass and excess-severity error, respectively.

\subsubsection{Continuous Ranked Probability Score (CRPS).} It is a strictly proper scoring rule to assess full-distribution performance that integrates squared differences between the predictive CDF and the realized-outcome step function~\cite{gneiting2007strictly}. 
\[
\operatorname{CRPS}(F,y)=\int_{-\infty}^{\infty}\{F(z)-\mathbf1\{y\le z\}\}^2\,dz.
\]
To emphasize outcomes above \(t\), we also use upper-tail threshold-weighted CRPS (twCRPS), which applies the weight \(w(x)=\mathbf1\{x\ge t\}\) to the same integrand~\cite{allen2024weighted,allen2023evaluating}. 
\[
\operatorname{twCRPS}(F,y)=\int_{-\infty}^{\infty}\{F(z)-\mathbf1\{y\le z\}\}^2w(z)\,dz
\]
Thus, CRPS evaluates the complete predictive distribution, whereas twCRPS concentrates the score on the operational upper-tail region. Both are negatively oriented scores, so lower values are better.

\section{Simulation Studies: Additional Details and Results}
\label{appdx:experiment_simulations}

This appendix documents the three controlled simulation studies summarized in Section~\ref{sec:experiments}. All studies use independently generated calibration and test data and evaluate fitted methods on a held-out set. Study (i) isolates occurrence and excess-severity distortions to verify the two-map construction and normalized-body preservation, Study (ii) gives a counterexample in which pooled PIT calibration is exact in population despite fixed-threshold severity miscalibration, and Study (iii) examines the finite-sample trade-off between correction and estimation variability as calibration exceedances become scarce. These designs are intended to characterize the operator and its target-specific trade-offs, rather than to establish superiority over global PIT recalibration.

\subsection{(i) Component-Wise Operator Verification}

This study verifies the occurrence--severity decomposition underlying PostTail in a setting where the two sources of fixed-threshold miscalibration can be controlled independently. At the fixed threshold \(t=0\), we specify an issued predictive CDF with heterogeneous exceedance probabilities, a subthreshold body, and a conditional upper tail. We then generate outcomes under four scenarios comprising occurrence miscalibration only, excess-severity miscalibration only, both distortions, and a population-calibrated identity case. This design checks that each fitted component addresses its intended purpose, that their recombination preserves normalized subthreshold shape, and that estimation can introduce finite-sample variation when no population correction is needed.

\subsubsection{Data Generating Process.}

At the fixed threshold \(t=0\), prediction \(i\) assigns the exceedance probability
\[
P_i=P_{\min}+(P_{\max}-P_{\min})V_i,
\qquad
P_{\min}=0.02,\quad P_{\max}=0.20,\quad
V_i\sim\operatorname{Beta}(2,5).
\]
Thus, \(P_i\) is the probability issued by the predictive CDF for the event \(Y_i>0\), and it varies across prediction instances from \(0.02\) to \(0.20\).  The lower bound ensures that every issued prediction has a nondegenerate upper tail, while the beta distribution concentrates the simulated predictions toward lower exceedance probabilities. These numerical values define a controlled design rather than application-specific tail probabilities. The mean issued exceedance probability is \(\mathbb E[P_i]=0.02+0.18(2/7)\approx0.0714\), which gives a reference scale for tail support. In scenarios with occurrence distortion, the true mean exceedance rate need not equal this value.
The issued predictive CDF is
\begin{equation}
F_i(y)=
\begin{cases}
0, & y\le -1,\\
(1-P_i)(y+1), & -1<y\le0,\\
1-P_i\exp(-y), & y>0.
\end{cases}
\label{eq:synthetic_data_generating}
\end{equation}
Conditional on \(Y_i\le0\), this CDF has a \(\operatorname{Unif}(-1,0)\) body. Conditional on \(Y_i>0\), it has exponential excess CDF \(G^+(y)=1-\exp(-y)\) for \(y>0\), whose inverse is \((G^+)^{-1}(z)=-\log(1-z)\). The uniform body and exponential tail are chosen for analytic simplicity. Their CDFs and inverse CDFs are explicit, the CDF joins continuously at the threshold, and occurrence and excess-severity distortions can be introduced independently. They provide a controlled verification setting and do not suggest that real predictive bodies or tails have these forms.

\subsubsection{Controlled Scenarios A to D.}
\label{appdx:scenarios_a_to_d}

Each scenario combines one of two occurrence laws with one of two excess-severity laws, yielding the four cases in Table~\ref{tab:synthetic-scenarios}.  For occurrence, the true exceedance probability is either the issued probability \(P_i\) or a smooth monotone distortion of it:
\[
\Pi_i=
\begin{cases}
P_i, & \text{occurrence calibrated},\\
\operatorname{logit}^{-1}\!\left\{0.5+0.8\operatorname{logit}(P_i)\right\},
    & \text{occurrence miscalibrated}.
\end{cases}
\qquad
E_i\mid P_i\sim\operatorname{Bernoulli}(\Pi_i),
\]
where \(E_i=\mathbf1\{Y_i>0\}\). The distorted law is representable by the beta occurrence-calibration family used by PostTail.\\ 
For severity, conditional on \(E_i=1\), the true excess-PIT is either uniform or drawn from a Beta distribution.
\[
Z_i\sim
\begin{cases}
\operatorname{Unif}(0,1), & \text{severity calibrated},\\
\operatorname{Beta}(2,1), & \text{severity miscalibrated}.
\end{cases}
\]
The \(\operatorname{Beta}(2,1)\) distribution places more excess-PIT values near one than the uniform calibration target, so generated exceedances are systematically more extreme than the issued conditional tail implies.
We generate the corresponding exceedance by applying the inverse issued-tail CDF, \(Y_i=(G^+)^{-1}(Z_i)=-\log(1-Z_i)\). Thus,
\[
\frac{F_i(Y_i)-F_i(0)}{1-F_i(0)}
=G^+(Y_i)
=1-\exp\{-[-\log(1-Z_i)]\}
=Z_i.
\]
That is, inverse-CDF generation deliberately makes \(Z_i\) the excess PIT of the generated exceedance under the issued prediction. When \(E_i=0\), we instead draw \(U_i\sim\operatorname{Unif}(0,1)\) and set \(Y_i=-1+U_i\). Scenarios A--D cross these occurrence and severity choices. Scenario A has occurrence error only, Scenario B has severity error only, Scenario C has both errors, and Scenario D is calibrated in population.

\begin{table}
\centering
\caption{Synthetic prediction scenarios and their intended diagnostics.}
\label{tab:synthetic-scenarios}
\begin{tabular}{llll}
\toprule
Scenario & Occurrence law & Excess-PIT law & Intended diagnostic \\
\midrule
A & Distorted $\Pi_i$
  & $\operatorname{Unif}(0,1)$
  & Occurrence correction with severity identity \\
B & $\Pi_i=P_i$
  & $\operatorname{Beta}(2,1)$
  & Occurrence identity with severity correction \\
C & Distorted $\Pi_i$
  & $\operatorname{Beta}(2,1)$
  & Both corrections required \\
D & $\Pi_i=P_i$
  & $\operatorname{Unif}(0,1)$
  & Population identity and finite-sample degradation \\
\bottomrule
\end{tabular}
\end{table}

Each scenario uses \(2{,}000\) calibration pairs and \(20{,}000\) independent test pairs. No training sample is needed because the base predictive CDF is analytic. All maps are fitted only on calibration outcomes and are frozen before test evaluation. Ten repetitions use seeds \(0,\ldots,9\). Scenario, split, and repetition identifiers define independent random-number streams, so the generated samples do not depend on execution order. Within each repetition, all methods use the same seed-specific calibration and test data, making their comparisons paired. Seed 0 is used only to select the illustrative component-ablation plot.

\subsubsection{Compared Methods.}

Table~\ref{tab:synthetic-compared-methods} summarizes the methods. Let
\(I(u)=u\) denote the identity map. The population-map variants are diagnostic
references. They use the data-generating maps and are therefore unavailable in
practice.
\begin{table}[t]
\centering
\caption{Methods in the component-wise simulation. O, S, and C denote
occurrence-only, severity-only, and combined PostTail, respectively.}
\label{tab:synthetic-compared-methods}
\begin{tabular}{llll}
\toprule
Method & Occurrence map & Severity map & Role \\
\midrule
Raw & \(I\) & \(I\) & Unmodified prediction \\
Global PIT (Kuleshov) &
\multicolumn{2}{l}{One isotonic map on the full PIT} &
Global recalibration comparator \\
\midrule
Population PT-O & \(r_{\mathrm{pop}}\) & \(I\) & Diagnostic reference \\
Population PT-S & \(I\) & \(s_{\mathrm{pop}}\) & Diagnostic reference \\
Population PT-C & \(r_{\mathrm{pop}}\) & \(s_{\mathrm{pop}}\) & Diagnostic reference \\
\midrule
Estimated PT-O & \(\widehat r\) & \(I\) & Occurrence ablation \\
Estimated PT-S & \(I\) & \(\widehat s\) & Severity ablation \\
Estimated PT-C & \(\widehat r\) & \(\widehat s\) & Combined PostTail \\
\bottomrule
\end{tabular}
\end{table}
Population maps are used to verify the implementation but are excluded from the estimated-method comparison because they require knowledge of the data-generating process. For diagnostic purposes, we also apply PostTail with the population
recalibration maps known from the data-generating process. The population
occurrence map is
\[
r_{\mathrm{pop}}(p)=
\begin{cases}
\operatorname{logit}^{-1}\!\left\{0.5+0.8\operatorname{logit}(p)\right\},
    & \text{in Scenarios A and C},\\
p, & \text{in Scenarios B and D},
\end{cases}
\]
and the population severity map is
\[
s_{\mathrm{pop}}(u)=
\begin{cases}
u, & \text{in Scenarios A and D},\\
u^2, & \text{in Scenarios B and C}.
\end{cases}
\]
Here, \(u^2\) is the CDF of the \(\operatorname{Beta}(2,1)\) excess-PIT
distribution used in the severity-miscalibrated scenarios. The population
PostTail curves apply the appropriate pair
\((r_{\mathrm{pop}},s_{\mathrm{pop}})\) for each scenario. They are not
implementable in practice, since the data-generating maps are unknown. Instead,
they distinguish the population-level effect of the operator from error due to
estimating the maps from finite calibration data.

\subsubsection{Results.} 
We first use component-wise ablations to distinguish population-map behavior from finite-sample estimation error, then summarize the held-out performance of the combined methods across repetitions. 

\begin{figure}
    \centering
    \includegraphics[width=0.95\linewidth]{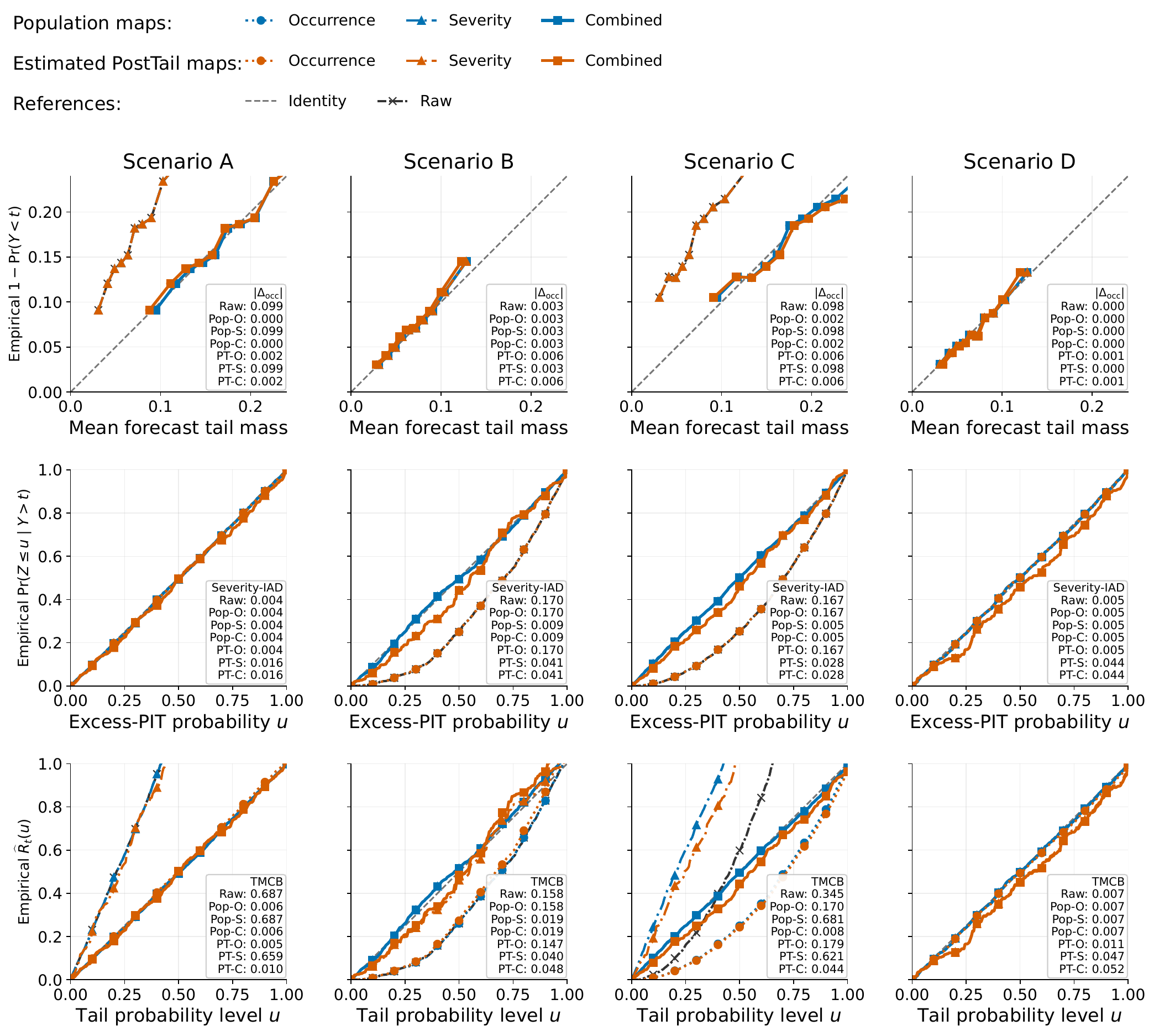}
    \caption{Population and estimated PostTail component ablations on a prespecified held-out repetition (seed 0). Columns show scenarios A--D. Rows show occurrence reliability, excess-PIT PP curves, and the combined finite-threshold calibration curve, respectively. In the top row, the horizontal coordinate is the bin-wise mean predicted tail mass, namely the predicted exceedance probability $\Pr_F(Y>t)$. Blue curves apply known population occurrence, severity, or both maps. Orange curves apply the corresponding calibration-sample estimates, and black curves are raw predictions. The population and estimated combined methods compose their occurrence and severity components. Insets report the held-out component metric for each plotted method. $|\Delta_{\mathrm{occ}}|$ is the absolute marginal occurrence gap, Excess/Severity-IAD is the integrated absolute deviation of the excess-PIT empirical CDF from the identity, and TMCB is the integrated absolute deviation of the combined curve. Pop and PT abbreviate population and estimated PostTail. O, S, and C abbreviate occurrence, severity, and combined. Coincident curves are expected when a component does not affect the diagnostic in that row. The combined-calibration row displays the non-redundant joint effect. Grey diagonals are identity calibration targets.}
    \label{fig:synthetic-population-estimated-ablations}
\end{figure}

Figure~\ref{fig:synthetic-population-estimated-ablations} compares raw predictions, population PostTail using these known maps, and estimated PostTail fitted from calibration data. It confirms the intended component separation on the prespecified representative repetition. In Scenario~A, the raw prediction has an occurrence gap but nearly uniform excess-PIT values. In Scenario~B, the reverse holds, and Scenario~C exhibits both errors. The corresponding population component maps remove only their targeted population distortion, while the combined population map corrects both. The estimated maps closely follow this behavior but retain finite-sample error. Scenario~D shows the complementary identity case, because neither component requires a population correction, fitting either map can introduce small deviations due to finite-sample variability.

\begin{figure}
    \centering
    \includegraphics[width=0.95\linewidth]{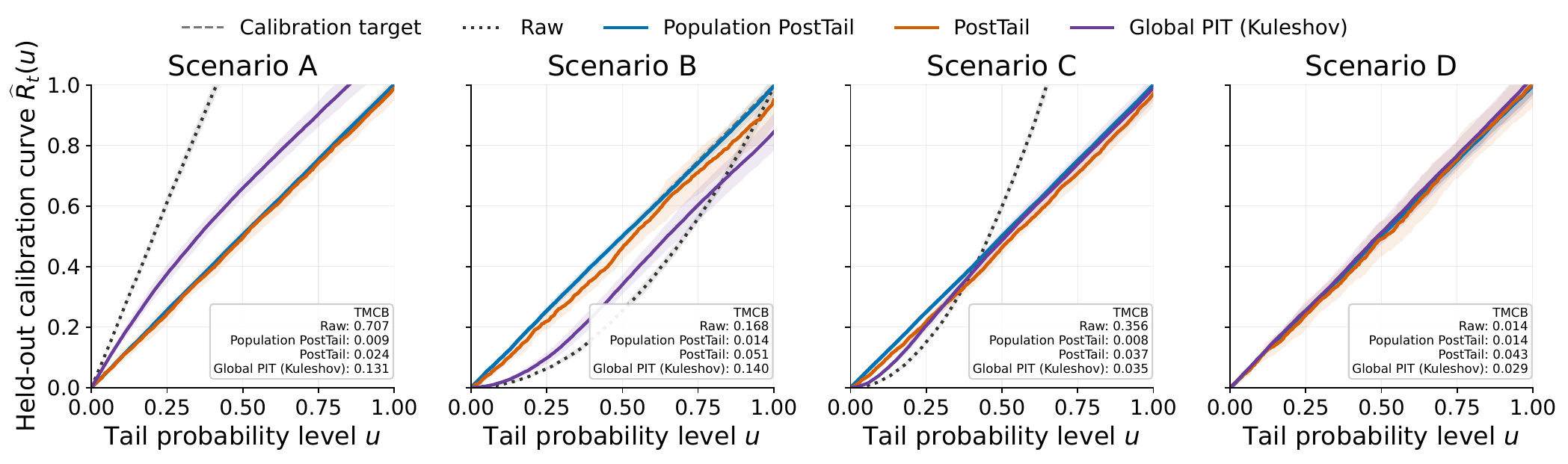}
    \caption{Held-out finite-threshold calibration curves across the four controlled scenarios. All curves and TMCB values are evaluated on independent test sets, after each estimated recalibrator is fitted only on its corresponding calibration set. Curves show the mean across 10 repetitions. Shaded bands show plus or minus one sample standard deviation across repetitions. Population PostTail applies the known occurrence and severity maps, PostTail estimates them from the calibration sample, and Global PIT applies Kuleshov isotonic recalibration to the complete predictive CDF. Insets report mean TMCB with lower values being better. The diagonal is the finite-threshold calibration target.}
    \label{fig:synthetic-tail-calibration-comparison}
\end{figure}

Figure~\ref{fig:synthetic-tail-calibration-comparison} summarizes this pattern across repetitions. For raw prediction, estimated PostTail, and global PIT, the mean TMCB values are \(0.707\), \(0.024\), and \(0.131\), respectively, in Scenario~A. The corresponding values are \(0.168\), \(0.051\), and \(0.140\) in Scenario~B, \(0.356\), \(0.037\), and \(0.035\) in Scenario~C, and \(0.014\), \(0.043\), and \(0.029\) in Scenario~D. These controlled studies verify that the combined operator addresses the designated component errors. Global PIT has lower mean held-out TMCB in Scenarios~C and~D, and fitted corrections increase TMCB in the calibrated identity case. These finite-sample comparisons complement the distinct question addressed next: whether pooled global-PIT recalibration can resolve a tail error when its population calibration target is already satisfied. More adaptive selection of the severity regularization weight may mitigate this finite-sample trade-off, but developing and validating such a selection rule requires future investigation. All reported A--D results use the fixed value \(\lambda_s=0.9\). To illustrate the distinction between the two calibration targets more directly, we next present a controlled counterexample in which global PIT is exactly calibrated in population but still has a fixed-threshold severity error that PostTail can correct.

\subsection{(ii) Global PIT Calibration with Tail Miscalibration}
\label{appdx:scenario_e_globally_calibrated_but_tail_miscalibrated}

\begin{figure}[t]
\centering
\includegraphics[width=0.95\linewidth]{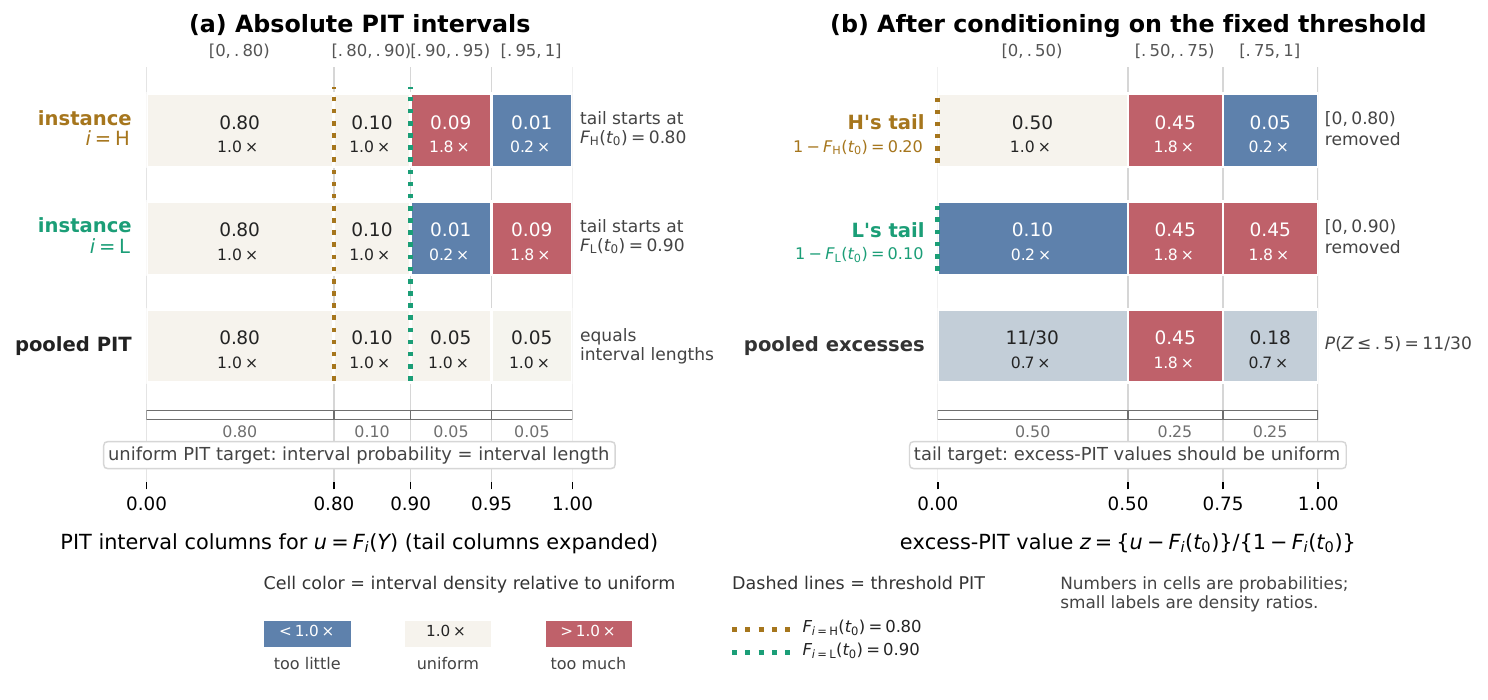}
\caption{
A globally uniform PIT can hide fixed-threshold severity error in Scenario E. (a) visualizes Table~\ref{tab:scenario-e-probabilities} for the two prediction types H and L.
Columns mark the displayed PIT intervals, with tail columns expanded for legibility. Cell numbers are conditional probabilities, and cell colors encode interval density relative to a uniform PIT distribution. The pooled row averages the two equally likely prediction types and exactly recovers the interval lengths, so the population global-PIT map is the identity. Panel (b) conditions on the fixed event \(Y>0\) and maps each prediction-specific tail to the aligned excess-PIT value \(Z=\{U-F_i(0)\}/\{1-F_i(0)\}\). On this scale, the pooled exceedance probabilities are not the uniform-tail probabilities, revealing excess-severity miscalibration that marginal global PIT cannot detect.}
\label{fig:scenario-e-interval-cancellation}
\end{figure}

This study is an extension of the previous scenarios and is a constructed toy example that highlights the difference between pooled global-PIT calibration and fixed-threshold tail calibration. With equal probability, an observation has prediction type \(S\in\{\mathrm H,\mathrm L\}\), with issued exceedance probabilities
\[
p_{\mathrm H}=0.20,\qquad p_{\mathrm L}=0.10.
\]
For both types, we use the analytic CDF family from Scenarios A--D, replacing
\(P_i\) by \(p_s\):
\[
F_s(y)=
\begin{cases}
0, & y\le -1,\\
(1-p_s)(y+1), & -1<y\le0,\\
1-p_s\exp(-y), & y>0.
\end{cases}
\]
Thus the common threshold \(t=0\) occurs at different locations on the PIT scale, \(F_{\mathrm H}(0)=0.80\) and \(F_{\mathrm L}(0)=0.90\). We construct outcomes directly on that scale. Conditional on \(S=s\), we draw \(U=F_s(Y)\) according to Table~\ref{tab:scenario-e-probabilities}, uniformly within each displayed interval, and set \(Y=F_s^{-1}(U)\). The table therefore specifies the conditional distribution of the raw PIT, and hence how the outcome distribution differs from the issued prediction. Figure~\ref{fig:scenario-e-interval-cancellation} visualizes this construction before and after each prediction-specific tail is normalized.

\begin{table}[t]
\centering
\caption{Conditional probabilities assigned to the four PIT intervals in Scenario E. The pooled row averages the two equally likely prediction types.}
\label{tab:scenario-e-probabilities}
\begin{tabular}{lcccc}
\toprule
Prediction type
& \([0,0.80)\)
& \([0.80,0.90)\)
& \([0.90,0.95)\)
& \([0.95,1]\) \\
\midrule
\(\mathrm H\) & \(0.80\) & \(0.10\) & \(0.09\) & \(0.01\) \\
\(\mathrm L\) & \(0.80\) & \(0.10\) & \(0.01\) & \(0.09\) \\
Pooled & \(0.80\) & \(0.10\) & \(0.05\) & \(0.05\) \\
\bottomrule
\end{tabular}
\end{table}

The first intuition is pooled cancellation. Because the two types are equally likely, averaging the two conditional rows gives interval masses \(0.80,0.10,0.05,0.05\), exactly the four interval lengths. Hence the pooled raw PIT satisfies \(U\sim\operatorname{Unif}(0,1)\) in population, and a global PIT recalibrator has no population-level correction to learn. This does not mean that each type is calibrated at every PIT level, rather their opposite distortions cancel only after pooling. The second intuition is that occurrence is correct within each type, whereas severity is not. Specifically,
\[
\Pr(Y>0\mid S=\mathrm H)=0.10+0.09+0.01=0.20=p_{\mathrm H},
\qquad
\Pr(Y>0\mid S=\mathrm L)=0.01+0.09=0.10=p_{\mathrm L}.
\]
After conditioning on exceedance, however, each type has a different tail
boundary. Writing
\[
Z_s=\frac{U-(1-p_s)}{p_s}\quad\text{on }Y>0,
\]
the type-\(\mathrm H\) masses over \([0,0.5)\), \([0.5,0.75)\), and
\([0.75,1]\) are \(0.50,0.45,0.05\), rather than the uniform values
\(0.50,0.25,0.25\). The type-\(\mathrm L\) masses over \([0,0.5)\) and
\([0.5,1]\) are \(0.10,0.90\), rather than \(0.50,0.50\). Panel~(b) of
Figure~\ref{fig:scenario-e-interval-cancellation} shows why this remains
visible after tail normalization. The interval \([0.80,0.90)\) is part of the
type-\(\mathrm H\) tail but remains in the type-\(\mathrm L\) body.
Among realized exceedances, types \(\mathrm H\) and \(\mathrm L\) have weights
\(2/3\) and \(1/3\), respectively, because their occurrence probabilities are
\(0.20\) and \(0.10\). Therefore,
\[
\Pr(Z\le0.5\mid Y>0)
=\frac23(0.50)+\frac13(0.10)
=\frac{11}{30}
\ne0.50.
\]
Therefore, this setup has exact population global-PIT calibration and correct type-specific occurrence, but fails excess-severity calibration at the fixed threshold. Heterogeneous threshold locations are essential. With one common threshold boundary, global PIT uniformity together with correct occurrence would also imply a uniform normalized tail.

\begin{figure}[t]
    \centering
    \includegraphics[width=0.95\linewidth]{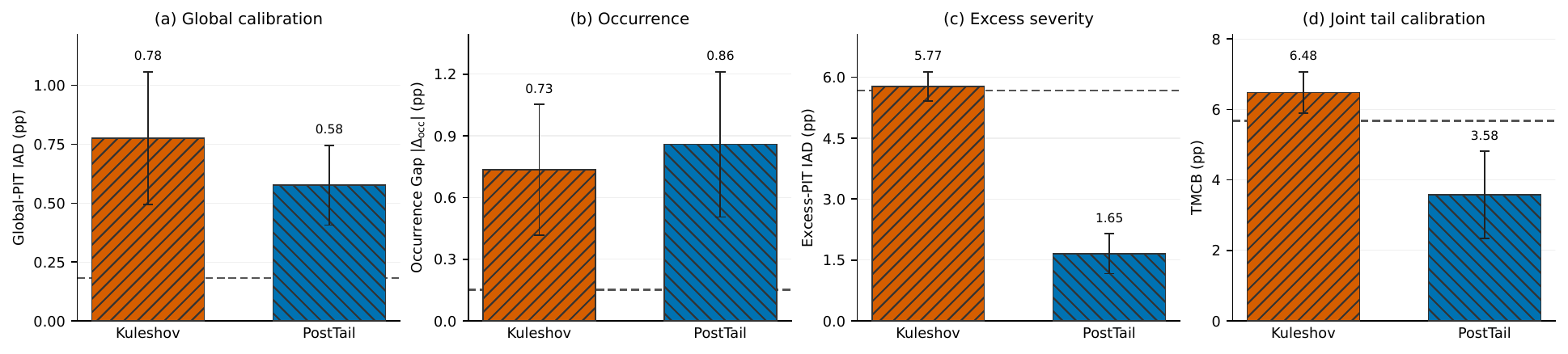}
    \caption{Comparison of Kuleshov isotonic global recalibration and combined PostTail. Panels report held-out global-PIT integrated absolute deviation (IAD), occurrence gap \(\left|\Delta_{\mathrm{occ}}\right|\), excess-PIT IAD, and finite-threshold calibration error (TMCB) across 10 repetitions. All metrics are multiplied by 100 and therefore represent average percentage-point deviations. Dashed horizontal lines mark the corresponding unmodified predictions, and error bars show $\pm 2$ standard errors across repetitions. Lower values are better.}
    \label{fig:appendix-scenario-e-metric-summary}
\end{figure}

Each of the ten repetitions uses \(2{,}000\) independent calibration pairs and \(20{,}000\) independent test pairs, as in Scenarios A--D. Kuleshov isotonic global PIT recalibration and combined PostTail are fitted only on the calibration pairs and evaluated only on the test pairs. Figure~\ref{fig:appendix-scenario-e-metric-summary} shows the resulting target-specific trade-off. PostTail lowers mean excess-PIT IAD from \(5.77\) to \(1.65\) percentage points and TMCB from \(6.48\) to \(3.58\) percentage points. Its global-PIT IAD remains small (\(0.58\) percentage points versus \(0.78\) for global PIT), but its occurrence gap is higher (\(0.86\) versus \(0.73\) percentage points). This controlled result illustrates a mismatch between pooled and fixed-threshold targets, rather than universal superiority of PostTail over global recalibration.

\subsection{(iii) Finite-Sample Behavior under Tail Scarcity}
\label{appdx:tail_sample_scarcity}

We run two complementary finite-sample stress tests. F1 isolates estimation of the severity map, whose effective sample size is the realized number \(m\) of calibration exceedances. F2 jointly estimates the occurrence and severity maps under the same scarcity grid. Both studies extend the fixed identity regularization in Equation~\eqref{eq:severity_regularization}. Rather than fixing the fitted-map weight, they use a prespecified weight that increases with \(m\). Thus, sparse tails favor no recalibration, whereas larger tail samples allow more of the fitted correction. These controlled studies characterize a bias--variance trade-off. They do not establish a universal safe tail-sample size.

\begin{table}[t]
\centering
\caption{Prespecified tail-scarcity design. \(m\) denotes the realized number
of calibration exceedances.}
\label{tab:tail-scarcity-design}
\begin{tabular}{ll}
\toprule
Factor & Values or construction \\
\midrule
Study & F1 severity only and F2 end-to-end PostTail \\
Mean prevalence & \(\pi_t\in\{0.01,0.05,0.10\}\) \\
Target tail support & \(\mathbb E[m]\in\{5,10,20,50,100,200\}\) \\
Calibration size & \(n=\operatorname{round}\{\mathbb E[m]/\pi_t\}\) \\
Error strength & Identity, moderate, or strong \\
Repetitions & 50 per simulation cell \\
Test support & 500 expected exceedances per cell \\
\bottomrule
\end{tabular}
\end{table}

Table~\ref{tab:tail-scarcity-design} gives the prespecified prevalence, tail-support, and repetition grid for both stress tests.
Seeds \(0,\ldots,49\) index the 50 Monte Carlo repetitions in every F1 and F2 simulation cell. They control the simulated calibration and test outcomes and hence the realized exceedance counts. Within each repetition, all methods use the same data, making their comparisons paired. The test draw is also reused across target-tail-count cells within a fixed study, prevalence, and error-strength setting to reduce Monte Carlo noise. Varying \(\pi_t\) and \(n\) while holding \(\mathbb E[m]\) fixed checks whether the realized tail count, rather than nominal calibration size alone, governs the estimation problem.
Studies (i) and (ii) use the fixed severity-map weight \(\lambda_s=0.9\), with occurrence-map weight one. Study (iii) compares unshrunk corrections with the prespecified count-adaptive shrinkage weight \(\lambda(m)=m/(m+50)\).

\subsubsection{F1: Controlled severity-map scarcity.}
It uses the continuous severity-only construction of Scenario B, in which occurrence probabilities are correct and the occurrence map remains the identity. The excess-PIT distribution is \(\operatorname{Beta}(1,1)\), \(\operatorname{Beta}(1.5,1)\), or \(\operatorname{Beta}(2,1)\), giving identity, moderate, and strong severity error, respectively. 
We compare raw predictions, the population map, ECDF, a smooth monotone Bernstein-polynomial mixture of degree \(8\) with roughness penalty \(0.1\), and Kuleshov isotonic severity maps. The Bernstein map provides a flexible continuous contrast to the discontinuous ECDF. For the ECDF map, we additionally evaluate count-adaptive identity shrinkage.
If \(m\ge1\), let \(\widehat s\) denote the fitted severity map and define
\[
\lambda(m)=\frac{m}{m+50},
\qquad
\widehat s_{\lambda}(u)
=(1-\lambda(m))u+\lambda(m)\widehat s(u).
\]
If \(m=0\), we use the identity map. The rule depends only on calibration counts and never on test outcomes. It assigns approximately \(9\%\), \(50\%\), and \(80\%\) weight to the fitted map at \(m=5\), \(50\), and \(200\), respectively.  The constant \(50\) is a prespecified half-weight scale for this stress test, not an optimal selector or universally recommended threshold.
\subsubsection{F2: End-to-end PostTail scarcity.}

It uses the combined-error construction of Scenario C, with the same prevalence, tail-support, and error-strength grid. It jointly fits a beta occurrence map and a smooth Bernstein severity map.
We compare the raw predictions, the population combined map, unshrunk and count-shrunk PostTail with beta occurrence and Bernstein severity, and Kuleshov global PIT recalibration.
The count-shrunk combined method applies the same realized-count weight to both fitted components.
\[
\widehat r_{\lambda}(p)
=(1-\lambda(m))p+\lambda(m)\widehat r(p),
\qquad
\widehat s_{\lambda}(u)
=(1-\lambda(m))u+\lambda(m)\widehat s(u).
\]
Because these are convex combinations with the identity, both maps remain monotone. Using \(m\) for both deliberately tests whether scarce tail support can provide a shared safeguard for the recombined operator, even though the occurrence map itself is fitted using all calibration pairs.

\subsubsection{Results.}

\begin{figure}[ht]
    \centering
    \includegraphics[width=0.95\linewidth]{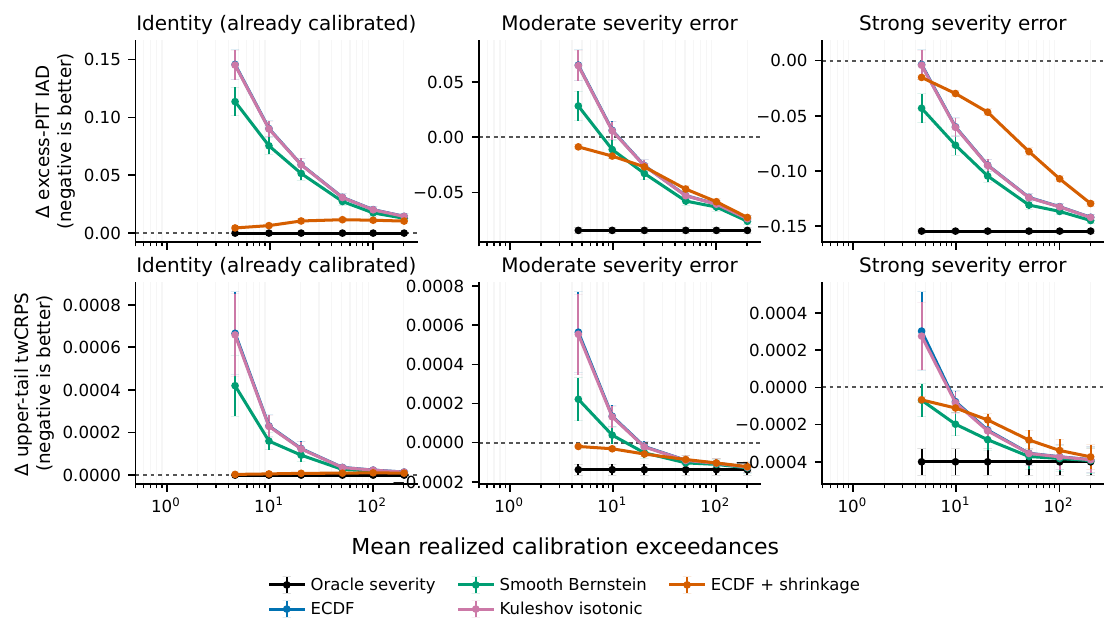}
    \caption{Finite-sample severity-map trade-off. Paired changes from the raw prediction in excess-PIT IAD and upper-tail twCRPS are shown against the mean realized number of calibration exceedances. The dotted horizontal line at zero denotes no change from the raw prediction. Error bars show $\pm 2$ standard errors across 50 repetitions. Negative values are better.}
    \label{fig:severity-scarcity-tradeoff}
\end{figure}

Figure~\ref{fig:severity-scarcity-tradeoff} shows that, under identity, the unshrunk flexible severity maps are harmful when \(m\) is about five. The ECDF, Bernstein, and Kuleshov maps increase excess-PIT IAD by roughly \(0.11\)--\(0.15\). 
Count-adaptive shrinkage reduces the mean increase for the ECDF map to about \(0.003\)--\(0.004\).
Under moderate error, the unshrunk maps remain harmful in the smallest cell, whereas the shrunk maps already improve severity IAD. By about \(m=20\), all estimators improve it. Under strong error, every estimator improves severity IAD from the smallest cell onward. The twCRPS panel shows a related but distinct trade-off. At the smallest counts, improved severity calibration can still coincide with worse upper-tail predictive quality.

\begin{figure}[h]
    \centering
    \includegraphics[width=0.95\linewidth]{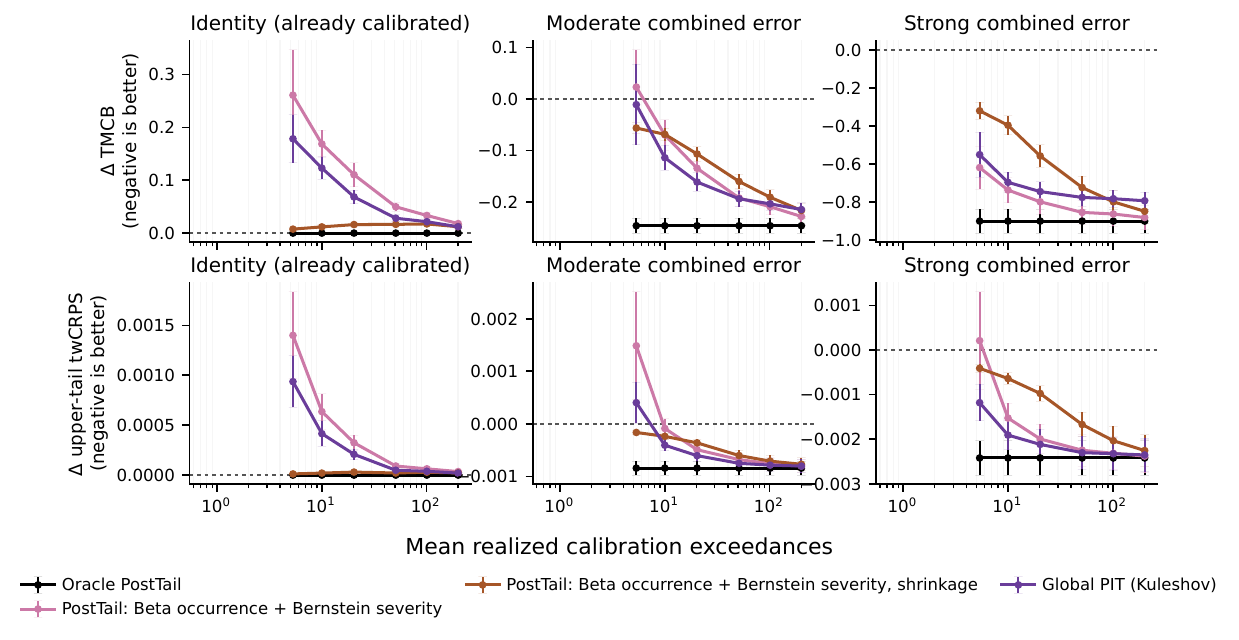}
    \caption{End-to-end PostTail under tail-sample scarcity. Paired changes from the raw prediction in finite-threshold calibration error (TMCB) and upper-tail twCRPS are shown against the mean realized number of calibration exceedances. The dotted horizontal line at zero denotes no change from the raw prediction. Error bars show $\pm 2$ standard errors across 50 repetitions. Negative values are better.}
    \label{fig:end-to-end-scarcity-tradeoff}
\end{figure}

Figure~\ref{fig:end-to-end-scarcity-tradeoff} shows the same pattern after joint estimation. In the identity cell near \(m=5\), unshrunk combined PostTail increases TMCB by \(0.261\), whereas count shrinkage reduces this to \(0.0075\). Under moderate error, shrinkage changes the smallest-cell TMCB difference from \(0.023\) to \(-0.056\). Under strong error, both versions improve TMCB, although only the shrunk version also improves twCRPS in that cell. At about \(m=200\), unshrunk combined PostTail gives the largest TMCB reduction (\(-0.881\)). These operating-region results do not identify a universal count threshold or an optimal shrinkage rule. Their useful range depends on the distortion, estimator, and evaluation metric.

Global PIT recalibration is included here to track finite-sample behavior as tail support increases, not to rank methods across all calibration targets. This particular toy construction can favor the global method on some metrics. That does not alter the purpose of the study, which is to expose the estimation behavior of the tail-specific maps under controlled scarcity. Simulation study (ii) provides the complementary construction in which pooled global PIT calibration is exact in population while fixed-threshold severity remains miscalibrated. Together, the two studies show that the methods have different targets, so this toy example does not establish the superiority or inferiority of global PIT recalibrators relative to PostTail.

\section{Real-Data Experiment: Additional Details and Results}
\label{appdx:experiment_realdata}

\subsection{Datasets, Splits, and Thresholds}

\begin{figure}[p]
    \centering
    \includegraphics[width=0.95\linewidth]{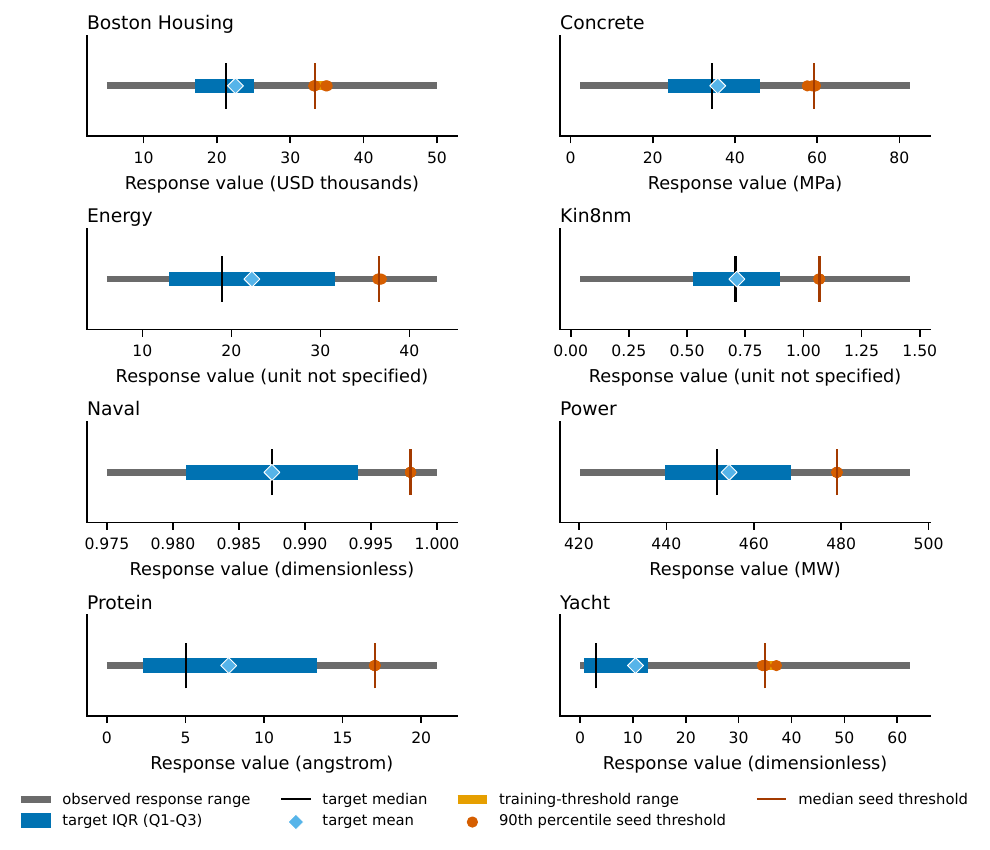}
    \caption{Each panel shows the observed response range (gray), its interquartile range (blue), median (black line), and mean (light-blue diamond), all on the response axis. The orange segment, points, and dark-orange line respectively show the range, individual values, and median of training 90th-percentile thresholds across seeds 0, 1, 2, 3, 4. Axes retain native response units and are not shared across datasets.}
    \label{fig:uci-threshold-context}
\end{figure}

For seeds \(0,\ldots,4\), we form training, calibration, and test splits of approximately \(72\%\), \(18\%\), and \(10\%\) of the dataset. Each seed controls the random permutation used for splitting as well as NumPy and PyTorch training randomness, including model initialization and minibatch order. Consequently, the training-only standardization, fitted model, calibration maps, and final test set vary by seed. The threshold \(t\) is the 90th percentile of that seed's training responses, which also produces slightly different held-out exceedance counts. Hyperparameter selection uses validation data and remains separate from final test reporting.
The event \(Y>t\) denotes the upper tail of the seed-specific training response distribution. It corresponds to high home value for Boston, high compressive strength for Concrete, high heating demand for Energy, a high kinematic output for Kin8nm, a high simulated propulsion-state coefficient for Naval, high power output for Power, a large protein-structure deviation for Protein, and high residuary resistance for Yacht. The datasets, their response variable and an illustrative use case for the high-tail threshold are provided in Table~\ref{tab:uci-threshold-context}. Corresponding threshold values in relation to the distribution of the response variable are depicted visually in Figure~\ref{fig:uci-threshold-context}. %

\begin{table}[t]
\centering
\small
    \begin{tabular}{p{0.12\textwidth}p{0.4\textwidth}p{0.42\textwidth}}
    \toprule
    Dataset & {\raggedright Response (unit)} & {\raggedright Illustrative use} \\
    \midrule
    Boston Housing
    & {\raggedright median owner-occupied home value (USD thousands)}
    & {\raggedright High-value appraisal or portfolio-risk screening} \\

    Concrete
    & {\raggedright concrete compressive strength (MPa)}
    & {\raggedright Strength acceptance or high-load design check} \\

    Energy
    & {\raggedright building heating load (unit not specified)}
    & {\raggedright Building-energy-demand assessment or efficiency screening} \\

    Kin8nm
    & {\raggedright robot end-effector distance from target (unit not specified)}
    & {\raggedright Large target-tracking-error monitoring} \\

    Naval
    & {\raggedright gas-turbine decay-state coefficient (dimensionless)}
    & {\raggedright Propulsion-condition monitoring} \\

    Power
    & {\raggedright hourly average net electrical power output (MW)}
    & {\raggedright Generation forecasting under varying ambient conditions} \\

    Protein
    & {\raggedright protein-structure RMSD (\AA{} - angstrom)}
    & {\raggedright Large-deviation structural-model validation triage} \\

    Yacht
    & {\raggedright residuary resistance per unit weight of displacement
    (dimensionless)}
    & {\raggedright High-resistance hull-design or propulsive-power check} \\
    \bottomrule
    \end{tabular}
    \caption{Response descriptions and illustrative operational uses for the
    regression benchmarks. The uses provide domain context only and are not
    claims about the benchmark data collection. The accompanying visualization
    summarizes the observed response ranges and seed-specific training thresholds.}
    \label{tab:uci-threshold-context}
\end{table}

The Wine dataset is also used for UCI regression, but it has only six observed response levels. For every seed, its training \(0.9\) quantile is rating \(7\), so the strict event \(Y>t\) is exactly the rare rating-\(8\) class. Two test splits and one calibration split contain no such observations. Treating this setting as continuous tail regression would therefore make the excess-severity comparison depend on an effectively empty ordinal class, so we omit it from our analysis.

\begin{table}[t]
\centering
\begin{tabular}{lcccc}
\toprule
Dataset & Calibration $n$ & Calibration $m$ & Test $n$ & Test $m$ \\
\midrule
Boston & 91 & 7--16 & 51 & 4--6 \\
Concrete & 186 & 12--25 & 103 & 5--11 \\
Energy & 139 & 11--22 & 77 & 6--12 \\
Kin8nm & 1475 & 130--155 & 820 & 76--95 \\
Naval & 2148 & 148--182 & 1194 & 86--106 \\
Power & 1723 & 153--183 & 957 & 89--113 \\
Protein & 8232 & 775--857 & 4573 & 403--476 \\
Yacht & 56 & 3--8 & 31 & 2--3 \\
\bottomrule
\end{tabular}
\caption{Calibration- and test-set sizes ($n$) and realized exceedance-count ranges ($m$) across random seeds. Counts vary because each seed changes both split membership and the training-split empirical $0.9$-quantile threshold.}
\label{tab:uci-tail-main-tail-support}
\end{table}

Table~\ref{tab:uci-tail-main-tail-support} reports the numbers of calibration and test instances for each dataset, together with the realized exceedance counts above the seed-specific threshold. Although the split sizes are fixed within a dataset, the exceedance counts vary across seeds because each seed yields a slightly different training set and hence a slightly different 90th-percentile threshold. Tail support ranges from only \(3\)--\(8\) calibration exceedances for Yacht to \(775\)--\(857\) for Protein.

\subsection{Model Training and Regularization Selection}

The MLP outputs \(250\) support points for each input, interpreted as an empirical predictive distribution. A logistic kernel converts this empirical distribution to a continuous predictive CDF. Its bandwidth is selected from \(\{0.01,0.025,0.05,0.1,0.2\}\) by CRPS on the validation set. The two-layer, width-64 architecture was chosen manually during preliminary development rather than through an architecture hyperparameter search, and was then fixed across datasets, seeds, and training objectives. All models use learning rate \(10^{-3}\) and zero weight decay.
The CRPS checkpoint is trained for \(300\) epochs. Each tail-aware candidate starts from that checkpoint and is fine-tuned for \(50\) epochs under its own objective, following the CRPS-pretraining and penalty-fine-tuning pattern of~\cite{wessel2026enforcing}. Thus, the \(300\)-epoch stage is performed once per dataset and seed. It is not repeated separately for all \(14\) tail-aware candidates. CRPS and CRPS+twCRPS use minibatches of size \(128\), whereas the TMCB penalty is evaluated on the full training set because it depends on the joint empirical tail-calibration curve.

Table~\ref{tab:uci-tail-main-loss-hyperparameters} lists the loss-specific candidate grids and validation selection rules. Neither coefficient has a universal scale because the relative magnitudes of CRPS, twCRPS, and TMCB depend on the model and data. The lower \(\gamma\) grid is therefore specific to this standardized-target setup. Final weights are selected using calibration data rather than test outcomes.
\begin{table}[!htbp]
\centering
\setlength{\tabcolsep}{3pt}
\begin{tabular}{p{0.15\linewidth}p{0.33\linewidth}p{0.4\linewidth}}
\toprule
Training loss & Candidate hyperparameter values & Validation selection rule \\
\midrule
CRPS & None & None \\
CRPS+\(\lambda\)twCRPS & \(\lambda\in\{0.1,0.25,0.5,0.75,1,1.5,2,3,4\}\) & Mean raw-model validation twCRPS, per dataset \\
CRPS+\(\gamma\)TMCB & \(\gamma\in\{0.01,0.03,0.1,0.3,1\}\) & Mean raw-model validation TMCB, per dataset \\
\bottomrule
\end{tabular}
\caption{Candidate hyperparameters for the loss functions. The twCRPS grid broadens the pilot grid \(\{0.5,1,2\}\) toward both weaker and stronger tail emphasis.}
\label{tab:uci-tail-main-loss-hyperparameters}
\end{table}
The full run contains one CRPS setting, nine CRPS+twCRPS settings, and five CRPS+TMCB settings for each dataset and seed. Across eight datasets and five seeds, this yields \(8\times5\times15=600\) fitted checkpoints before validation selection. The selected \((\lambda,\gamma)\) pairs are Boston \((4,0.3)\), Concrete \((4,0.03)\), Energy \((4,0.01)\), Kin8nm \((1,0.03)\), Naval \((2,0.1)\), Power \((4,0.01)\), Protein \((4,0.01)\), and Yacht \((3,1)\). %

\FloatBarrier
\subsection{Additional Results}
In this section we report aggregate post-hoc performance and dataset-wise paired TMCB comparisons to complement the main real-data results. The real-data PostTail experiments use \(\lambda_r = 1.0\) and \(\lambda_s = 0.9\).

\begin{table}[!htbp]
\centering
\setlength{\tabcolsep}{3pt}
\begin{tabular}{lcccccc}
\toprule
Post-hoc & Rel. CRPS $\downarrow$ & Rel. twCRPS $\downarrow$ & $|\Delta_{\mathrm{occ}}| \downarrow$ & Excess-IAD $\downarrow$ & TMCB $\downarrow$ & Global-IAD $\downarrow$ \\
\midrule
Raw (Uncalibrated) & $0.998$ & $0.979$ & $1.3$ & $12.8$ & $16.3$ & $5.0$ \\
Global PIT & $0.985$ & $0.978$ & $1.2$ & $12.6$ & $16.4$ & $3.2$ \\
PostTail (ours) & $0.998$ & $0.965$ & $0.9$ & $10.4$ & $13.3$ & $4.7$ \\
\bottomrule
\end{tabular}
\caption{Post-hoc comparison across $n=8$ equally weighted UCI datasets and 3 selected training-loss families. Each entry first averages seeds within a dataset and training family, then averages all dataset--family cells equally. CRPS and twCRPS are ratios to the raw CRPS-trained model within each dataset and seed. Calibration metrics are shown $\times 100$. $\lambda$ for CRPS+$\lambda$twCRPS is selected per dataset by mean raw-model validation based on twCRPS. $\gamma$ for CRPS+$\gamma$TMCB is selected analogously by validation TMCB.}
\label{tab:uci-tail-main-posthoc-average}
\end{table}
Table~\ref{tab:uci-tail-main-posthoc-average} summarizes the post-hoc methods across all datasets and training losses. PostTail attains the lowest aggregate fixed-threshold errors. Its occurrence gap, Excess-IAD, and TMCB are \(0.9\), \(10.4\), and \(13.3\) percentage points, respectively, compared with \(1.3\), \(12.8\), and \(16.3\) for the unmodified predictions. Its aggregate relative CRPS matches that of the unmodified predictions at the reported precision (\(0.998\)), while relative twCRPS decreases from \(0.979\) to \(0.965\). These averages do not imply improvements for every dataset, seed, or training family. Global PIT instead achieves the lowest Global-IAD (\(3.2\) versus \(4.7\) for PostTail), consistent with its pooled-PIT target. Thus, the aggregate results distinguish improved fixed-threshold calibration from improved pooled global-PIT calibration rather than indicating uniform dominance by either recalibrator.
PostTail changes the total tail mass and the conditional shape above the fixed threshold while preserving the normalized subthreshold shape. Because CRPS averages discrepancy over the full predictive CDF and twCRPS still averages over the entire upper-tail region, these localized, generally modest corrections can substantially improve calibration diagnostics without greatly changing either proper score.

\begin{table}[!htbp]
\centering
\begin{tabular}{lccc}
\toprule
Dataset & Global PIT & CRPS+$\lambda$twCRPS & CRPS+$\gamma$TMCB \\
\midrule
Boston & 10/15 (66.7\%) & 4/5 (80.0\%) & 3/5 (60.0\%) \\
Concrete & 7/15 (46.7\%) & 2/5 (40.0\%) & 2/5 (40.0\%) \\
Energy & 6/15 (40.0\%) & 3/5 (60.0\%) & 3/5 (60.0\%) \\
Kin8nm & 8/15 (53.3\%) & 2/5 (40.0\%) & 2/5 (40.0\%) \\
Naval & 15/15 (100.0\%) & 5/5 (100.0\%) & 5/5 (100.0\%) \\
Power & 11/15 (73.3\%) & 4/5 (80.0\%) & 2/5 (40.0\%) \\
Protein & 14/15 (93.3\%) & 4/5 (80.0\%) & 4/5 (80.0\%) \\
Yacht & 9/15 (60.0\%) & 3/5 (60.0\%) & 2/5 (40.0\%) \\
\bottomrule
\end{tabular}
\caption{Dataset-wise TMCB wins for PostTail. 
Counts summarize paired comparisons across five seed runs per dataset. Each Global PIT cell counts strict wins over 15 seed--training-family comparisons. The two training-only tail-aware columns are compared to PostTail applied to the CRPS-trained model and use five paired seeds per dataset.
Ties count as non-wins. CRPS+twCRPS $\lambda$ is selected per dataset by mean raw-model validation twCRPS. CRPS+TMCB $\gamma$ is selected analogously by validation TMCB.}
\label{tab:uci-tail-main-posttail-tmcb-win-rates-by-dataset}
\end{table}
Table~\ref{tab:uci-tail-main-posttail-tmcb-win-rates-by-dataset} shows PostTail's TMCB advantage by dataset. Against global PIT recalibration, PostTail wins a majority of the paired comparisons on six of the eight datasets, with all \(15/15\) wins on Naval and \(14/15\) on Protein, but fewer than half on Concrete and Energy. The comparisons with tail-aware training are similarly mixed, although PostTail wins all five paired comparisons on Naval against both training objectives. These dataset-wise paired counts complement the aggregate results by showing that the fixed-threshold advantage is not uniform across benchmark regimes.

\subsection{Hardware and Software Used}
\label{appdx:experiment_reproducibility}

All simulation and real-data training, calibration, evaluation, and analysis were run on the HP EliteBook 840 G5 laptop with an Intel Core i5-8250U CPU. 
All computations used the CPU with no GPU involved.
We used a Python 3.10.16 environment with PyTorch 2.10.0+cu128, NumPy 2.2.5, Pandas 2.3.3, Matplotlib 3.10.8, and betacal 1.1.0. %

\FloatBarrier

\end{document}